\documentclass[lettersize,journal]{IEEEtran}
\PassOptionsToPackage{table}{xcolor}
\usepackage[edges]{forest}
\usepackage{amsmath,amsfonts,amssymb}
\usepackage{newtxtext,newtxmath}
\usepackage{array}
\usepackage{booktabs,tabularx}
\usepackage{longtable}
\usepackage{placeins}
\usepackage[table]{xcolor}
\usepackage{stfloats}
\usepackage{pifont}
\usepackage{url}
\usepackage{graphicx}
\usepackage{cite}
\usepackage{tikz}
\usepackage{multirow}
\usetikzlibrary{calc}
\makeatletter
\def\fps@figure{t}
\def\fps@table{t}
\makeatother
\newcommand{\SurveyFigureCaption}[2]{%
  \caption[#1]{\textbf{#1.} #2}}
\input{links}
\definecolor{SurveyTableSection}{HTML}{EAE3F3}
\definecolor{SurveyTableHeader}{HTML}{D8CDE6}
\definecolor{SurveyTableStripe}{HTML}{FAF9FC}
\definecolor{SurveyTableRule}{HTML}{B8AEC7}
\newcommand{\SurveyTableFont}{\fontsize{7}{8.4}\selectfont}
\newcommand{\SurveyDescriptionFont}{\fontsize{7.5}{9}\selectfont}
\newcommand{\SurveyTableCaptionFont}{\SurveyDescriptionFont}
\newcommand{\SurveyTableRowColor}[1]{\rowcolor{#1}[\tabcolsep][\tabcolsep]}
\newcommand{\SurveyTableStyle}{%
  \let\footnotesize\SurveyTableFont
  \let\scriptsize\SurveyTableFont
  \SurveyTableFont\color{black}%
  \hyphenpenalty=250\exhyphenpenalty=250
  \renewcommand{\arraystretch}{1.24}%
  \setlength{\tabcolsep}{1.5pt}%
  \setlength{\extrarowheight}{.2pt}%
  \setlength{\aboverulesep}{0pt}%
  \setlength{\belowrulesep}{0pt}%
  \setlength{\abovetopsep}{0pt}%
  \setlength{\belowbottomsep}{0pt}%
  \setlength{\arrayrulewidth}{.35pt}%
  \setlength{\lightrulewidth}{.35pt}%
  \arrayrulecolor{SurveyTableRule}}
\newcommand{\SurveyTableCaption}[2]{%
  \begingroup
  \let\footnotesize\SurveyTableCaptionFont
  \caption{#1}\label{#2}%
  \endgroup}
\newcommand{\SurveyTableDescription}[1]{%
  \par\vspace{2pt}%
  \noindent\begin{minipage}{\linewidth}\SurveyDescriptionFont\raggedright
    #1%
  \end{minipage}}
\newcommand{\SurveyTablePanel}[2]{%
  \SurveyTableRowColor{SurveyTableSection}%
  \multicolumn{#1}{l}{\rule[-1.5pt]{0pt}{11pt}\textbf{#2}} \\}
\newcommand{\SurveyTableTopRule}{\toprule[.8pt]}
\newcommand{\SurveyTableHeaderRule}{\midrule[.4pt]}
\newcommand{\SurveyTableBottomRule}{\bottomrule[.8pt]}

\begin{document}

\bstctlcite{IEEEexample:BSTcontrol}

\title{Vision in the Era of Agents: Toward Recursive Self-Improvement}

\author{Dong Chen, Shuai Ge, Yu Fu, Xiangyu Ren, Zhihao Zhang, Hao Wang, Changshuo Wang, \\Jiayi Ji, Henghui Ding, and Shuting He

\thanks{D. Chen, S. Ge, Y. Fu, X. Ren, Z. Zhang, and S. He are with Shanghai University of Finance and Economics, China. (e-mail: shuting.he@sufe.edu.cn).}
\thanks{H. Wang is with Mohamed bin Zayed University of Artificial Intelligence, United Arab Emirates.}%
\thanks{C. Wang is with University College London, United Kingdom.}
\thanks{J. Ji is with Xiamen University, China.}
\thanks{H. Ding is with Fudan University, China.}}
\markboth{}
{Shell \MakeLowercase{\textit{et al.}}: A Sample Article Using IEEEtran.cls for IEEE Journals}

\maketitle

\begin{abstract}
Large language model (LLM) agents have emerged as a paradigm for goal-directed problem solving, leveraging reasoning, planning, and tool use to accomplish complex tasks. Beyond text-based applications, these capabilities are increasingly being extended to visual domains, where visual agents ground their decisions and actions in images, videos, and interactive feedback. This paper provides a systematic overview of recent advances in these visual agents. We first conceptualize visual agents as systems in which one or more models operate within an associated agent harness. We then categorize mechanisms for improving agent capabilities by the depth of experience-driven change in these systems, distinguishing within-task adaptation, self-evolution, and recursive self-improvement (RSI). For each level, we review representative methods across visual understanding, generation, computer use, and embodied interaction, examining their sources of experience, targets of change, and underlying mechanisms. Relevant benchmarks and evaluation protocols are also summarized, providing a reference for evaluating visual agents across diverse tasks and settings. To support further research, resources are available at \url{http://visual-agent-survey.github.io}.
\end{abstract}


\section{Introduction}
\label{sec:introduction}

In an emerging problem-solving paradigm, large language model (LLM) agents pursue goals by combining reasoning, planning, and tool use~\cite{ICLR2023ReAct,ICLR2025AgentSquare}. They extend the capabilities of LLMs beyond text generation by interacting with external environments~\cite{zhang2026generalizability,NeurIPS2024AVATAR}. Observations and feedback from these interactions guide subsequent decisions, allowing agents to gather information, check intermediate results, and revise their plans. Agents can also use experience from earlier attempts to improve their ability to solve tasks~\cite{NeurIPS2023Reflexion,wang2023voyager,ICLR2026Darwin}. These capabilities support applications in information seeking~\cite{EMNLP2025DeepResearcher}, software development~\cite{ICLR2025OpenHands}, and mathematical problem solving~\cite{ICLR2026AgentMath}.

\emph{Visual agents} extend this paradigm to visual perception and interaction tasks, including visual understanding~\cite{CVPR2023Visual,ICCV2023ViperGPT,NeurIPS2025Deep}, image and video generation and editing~\cite{NeurIPS2024GenArtist,AAAI2026Genmac,EMNLP2026VideoAgentEditing}, GUI perception and grounding~\cite{NeurIPS2025GUIActor}, computer use in web and desktop environments~\cite{ICLR2024WebArena,NeurIPS2024OSWorld,agashe2025agent}, and embodied interaction~\cite{ICML2019PlaNet,ICLR2020Dreamer,added_wang2025jarvisone}. These tasks require agents to extract task-relevant information from visual inputs and reason about spatial and temporal relationships. Compared with agents relying mainly on textual observations, visual agents ground their decisions and actions in images, videos, and visual feedback, enabling interaction with visually changing environments~\cite{xie2025large,mokaria2026survey}. This integration enables agents to retrieve relevant visual evidence, inspect generated outputs, and select actions based on the observed environment state. Beyond fixed-workflow execution, recent work explores how output and action feedback can support self-evolution and recursive self-improvement (RSI)~\cite{added_sun2026seagent,ICML2026Agent0VL}.

The rapid development of visual agents and their expanding range of applications have motivated a growing body of work reviewing recent progress. While existing reviews of multimodal agents cover architectures, capabilities, and applications~\cite{xie2025large,mokaria2026survey}, they do not focus on self-evolution or RSI. In contrast, existing work on self-evolution and RSI primarily targets text-based or general-purpose agents, treating visual agents as at most one component among many~\cite{ren2026selfimprovements,xu2026evolvinggraphs,liu2026path,yu2026harness,jiang2026crossscenario}. Meanwhile, domain-specific reviews examine visual agents in areas such as visual generation, computer use, and embodied interaction~\cite{xing2026agenticvisualgeneration,huang2026agentic,fan2026video,sager2026comprehensive,gu2022vlnsurvey,qin2026towards,luo2026toward,han2026recursivephysical}, but each addresses only a narrow subset of the field rather than providing a systematic classification of how visual agents as a whole adapt and improve. These gaps motivate a dedicated and unified view of the mechanisms through which visual agents adapt and improve.

In this work, we view visual agents as systems comprising one or more models operating within an agent harness. Building on this view, we distinguish within-task adaptation, self-evolution, and RSI by the target, persistence, and subsequent use of experience-driven changes. We use this taxonomy to synthesize research on visual understanding, generation and editing, computer use, and embodied interaction. We also review benchmarks and evaluation protocols, discuss open challenges, and outline directions for future research.

\begin{figure*}[!t]
    \centering
    \input{figures/taxonomy}
    \vspace{-6mm}
    \SurveyFigureCaption{Taxonomy of visual agent adaptation and evolution}{The three levels cover within-task adaptation (L1), self-evolution (L2), and work toward recursive self-improvement (L3), as discussed in Sections~\ref{sec:level1}--\ref{sec:level3}. Each branch identifies a mechanism subcategory, gives its subsection number, and lists representative works.}
    \vspace{-2mm}
    \label{fig:taxonomy}
\end{figure*}

\noindent$\bullet$\ \textbf{Contributions.} This paper provides a structured overview of research on visual agent adaptation and evolution. We first clarify the complementary roles of model capabilities and agent harnesses to establish a common conceptual foundation. Building on this foundation, we organize the main body of the paper around a taxonomy that distinguishes within-task adaptation, self-evolution, and RSI. For each category, we discuss representative methods across different visual domains. We further examine existing benchmarks and evaluation protocols to clarify the extent to which their results support claims of improvement. Finally, based on this analysis, we identify open challenges and outline promising directions for future research on visual agent adaptation and evolution.

\noindent$\bullet$\ \textbf{Organization.} Fig.~\ref{fig:taxonomy} presents an overview of the taxonomy that structures this paper. Section~\ref{sec:background} introduces the background of visual agents, including their definition and scope. It also discusses the complementary roles of models and agent harnesses and establishes the taxonomy of adaptation and evolution. Sections~\ref{sec:level1}, \ref{sec:level2}, and~\ref{sec:level3} examine existing research on within-task adaptation, self-evolution, and emerging approaches to RSI, respectively. Section~\ref{sec:benchmarks} examines benchmarks and evaluation protocols for assessing these mechanisms. Section~\ref{sec:future-directions} discusses open challenges and outlines promising directions for future research. Finally, Section~\ref{sec:conclusion} concludes the paper by summarizing its main insights.

\section{Background}
\label{sec:background}

\subsection{Definition and Scope}
\label{sec:definition-scope}

\noindent$\bullet$\ \textbf{Definition.} In this survey, a \emph{visual agent} is a goal-directed system that uses visual observations to guide the selection and execution of task-level operations. These operations may include acquiring evidence, invoking tools, generating or editing visual content, and interacting with digital or physical environments. Its outputs may include responses grounded in visual evidence, visual artifacts, or actions in an environment.

\noindent$\bullet$\ \textbf{Scope.} We cover visual agents for visual understanding and for generating and editing visual content. Our review also extends to emerging interactive applications, including embodied interaction and computer use ranging from graphical user interface (GUI) perception and grounding to end-to-end workflows. We examine how an agent's own experience and external inputs such as human demonstrations and evaluation feedback can inform changes to task execution, agent resources, and improvement mechanisms.

\subsection{Foundations of Visual Agents}
\label{sec:visual-agent-foundations}

\begin{figure*}[!t]
  \centering
  \includegraphics[width=\textwidth]{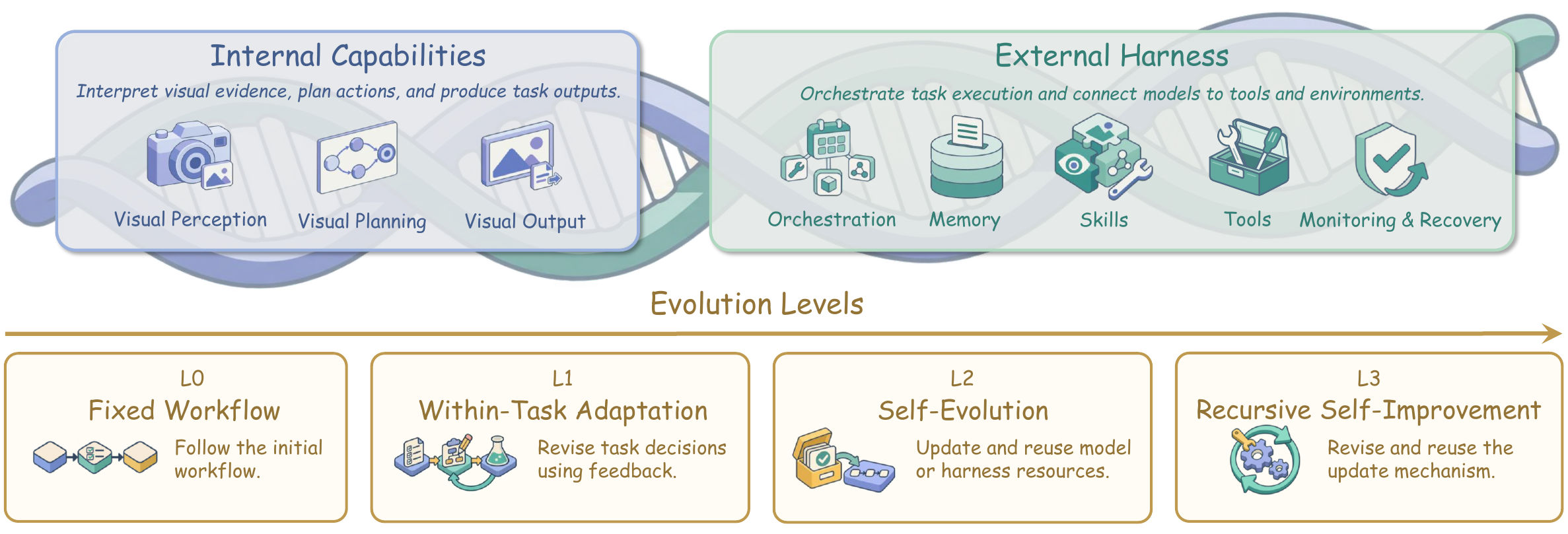}
  \vspace{-6mm}
  \SurveyFigureCaption{Visual agent foundations and levels of adaptation and evolution}{%
    Models provide visual capabilities, and the harness orchestrates execution.
    The lower row shows a fixed workflow (L0), within-task adaptation from feedback (L1), self-evolution through persistent resource updates by a fixed updater (L2), and recursive self-improvement (RSI) through reuse of a revised updater (L3).}
  \label{fig:foundations-dna}
  \vspace{-2mm}
\end{figure*}

A visual agent combines one or more models with an agent harness. Models provide internal capabilities for visual perception, planning, and output. The harness integrates orchestration, memory, reusable skills, tools, and monitoring and recovery mechanisms to coordinate execution and connect the agent to its environment~\cite{xie2025large,mokaria2026survey}. Training can improve model capabilities~\cite{CVPR2026AtomicVLA,ICLR2026Genie,ICML2026CoEvolving} and optimize harness policies for functions such as task scheduling and skill library maintenance~\cite{deng2026training,bai2026practice}. We write the agent state at evolution stage $k$ as
\begin{equation}
  \mathcal{A}_k = \bigl(
  \underbrace{\mathcal{M}_{\theta_k}}_{\text{models}},
  \underbrace{\mathcal{H}_{\eta_k}}_{\text{harness}}
  \bigr),
  \label{eq:agent-foundations}
\end{equation}
where $\theta_k$ denotes the learned model parameters and $\eta_k$ represents the configuration and resources of the harness. This formulation distinguishes updates to learned model capabilities from those to the external resources and procedures that support their use. Fig.~\ref{fig:foundations-dna} illustrates these complementary parts.

At execution step $t$, the agent combines its model component with its harness to act on the current task state. We formalize this process as
\begin{equation}
  (a_t, z_{t+1}) =
  \mathcal{H}_{\eta_k}[\mathcal{M}_{\theta_k}](g, o_t, z_t, W_t).
  \label{eq:visual-agent-execution}
\end{equation}
Here, $g$ denotes the task goal, and $o_t$ represents the current visual observation. The execution strategy $W_t$ is the active workflow for the current task. It specifies the selection and order of steps for evidence acquisition, planning, tool use, and control. The transient working state $z_t$ records workflow progress and plan details for execution, such as grounded targets and operation arguments. The outputs $a_t$ and $z_{t+1}$ denote the operation executed at step $t$ and the resulting working state, respectively. The operator $\mathcal{H}_{\eta_k}[\mathcal{M}_{\theta_k}]$ represents the agent's execution process as implemented by the model component and harness. The harness supplies context to model calls, invokes tools as needed, executes the selected operations, and incorporates outcome feedback into subsequent steps.

Before execution, the initial workflow $W_0$ is selected or instantiated based on the task goal and initial observations. We use fixed-workflow execution as the Level~0 (L0) reference, illustrated in Fig.~\ref{fig:level-characteristics}(a). In this setting, the agent follows its initial workflow throughout the task~\cite{CVPR2023Visual,ICCV2023ViperGPT}, so the active workflow at each execution step $t$ satisfies
\begin{equation}
  W_t = W_0.
  \label{eq:l0-fixed-workflow}
\end{equation}
Within this fixed workflow, the current visual observation $o_t$ supplies input to the scheduled operation, and the working state $z_t$ records execution progress. At the task-level, execution feedback does not revise current decisions or proposed outputs.

\subsection{Taxonomy}
\label{sec:taxonomy}

Building on the L0 reference, our taxonomy defines Level~1 (L1) as within-task adaptation, Level~2 (L2) as self-evolution, and Level~3 (L3) as recursive self-improvement (RSI).

The index $t$ tracks execution steps within a task, and $k$ tracks stages of agent evolution. In Fig.~\ref{fig:level-characteristics}(a) and (b), $\mathcal{A}_0$ denotes the models and harness used to execute the illustrated workflows. At L1, the agent can revise the active workflow $W_t$ in response to task feedback. Experience guides updates to model or harness resources at L2 and revisions to update mechanisms at L3. We distinguish these processes by what changes, whether the change persists, and how it is used later.

\begin{figure}[!t]
  \centering
\begingroup
\definecolor{LevelCFixed}{HTML}{596579}
\definecolor{LevelCOutline}{HTML}{426578}
\colorlet{LevelCAdapt}{LevelCOutline}
\colorlet{LevelCEvolve}{LevelCOutline}
\colorlet{LevelCRecursive}{LevelCOutline}
\definecolor{LevelCAdaptFill}{HTML}{FAF4DD}
\definecolor{LevelCEvolveFill}{HTML}{F8E6E6}
\definecolor{LevelCRecursiveFill}{HTML}{E6EEF8}
\colorlet{LevelCAdaptStroke}{LevelCAdapt}
\colorlet{LevelCEvolveStroke}{LevelCEvolve}
\colorlet{LevelCRecursiveStroke}{LevelCRecursive}
\colorlet{LevelCFixedFill}{LevelCFixed!4}
\colorlet{LevelCFixedStroke}{LevelCOutline}
\colorlet{LevelCFixedHeader}{LevelCFixed!10!white}
\colorlet{LevelCAdaptHeader}{LevelCAdaptFill}
\colorlet{LevelCEvolveHeader}{LevelCEvolveFill}
\colorlet{LevelCRecursiveHeader}{LevelCRecursiveFill}
\newlength{\LevelCCardWidth}
\newlength{\LevelCTopHeight}
\newlength{\LevelCBottomHeight}
\newlength{\LevelCColumnGap}
\newlength{\LevelCRowGap}
\newlength{\LevelCResourceWidth}
\setlength{\LevelCColumnGap}{8pt}
\setlength{\LevelCRowGap}{8pt}
\setlength{\LevelCTopHeight}{88pt}
\setlength{\LevelCBottomHeight}{102pt}
\setlength{\LevelCResourceWidth}{38pt}
\setlength{\LevelCCardWidth}{\dimexpr(\linewidth-\LevelCColumnGap-.6pt)/2\relax}
\newcommand{\LevelCPanel}[6]{%
  \begin{scope}[
      xshift=\dimexpr#1\LevelCCardWidth+#1\LevelCColumnGap\relax,
      yshift=\dimexpr-#2\LevelCTopHeight-#2\LevelCRowGap\relax]
    \path[fill=#3Fill!65!white,rounded corners=3pt]
    (0,0) rectangle (\LevelCCardWidth,-#6);
    \path[fill=#3Header,rounded corners=3pt]
    (0,0) rectangle (\LevelCCardWidth,-24pt);
    \draw[draw=#3Stroke,line width=.55pt,rounded corners=3pt]
    (0,0) rectangle (\LevelCCardWidth,-#6);
    \node[font=\bfseries\fontsize{8.5}{9.5}\selectfont]
    at (.5\LevelCCardWidth,-7.5pt) {#4};
    \node[font=\bfseries\fontsize{7.3}{8.5}\selectfont]
    at (.5\LevelCCardWidth,-18.5pt) {#5};
  \end{scope}}
\newcommand{\LevelCAgentFrame}{%
  \draw[draw=LevelCOutline!70!white,line width=.45pt,rounded corners=2pt]
  (4pt,-27pt) rectangle (\dimexpr\LevelCCardWidth-4pt\relax,-83pt);
  \node[levelC note,anchor=west,text=LevelCOutline]
  at (8pt,-33.5pt) {$\mathcal{A}_0$};}
\begin{tikzpicture}[x=1pt,y=1pt,
    levelC mini/.style={draw=LevelCOutline,line width=.45pt,
        fill=LevelCFixedFill!35!white,rounded corners=2pt,
        minimum height=15pt,inner xsep=2pt,inner ysep=1pt,
        font=\fontsize{7.2}{8.2}\selectfont},
    levelC step/.style={levelC mini,minimum width=28pt},
    levelC symbol/.style={levelC mini,minimum width=22pt},
    levelC arrow/.style={->,>=stealth,line width=.65pt,draw=black!65},
    levelC note/.style={font=\fontsize{7}{8}\selectfont,text=black!78,
        inner sep=1pt}]
  \LevelCPanel{0}{0}{LevelCFixed}{(a) L0}{Fixed Workflow}{\LevelCTopHeight}
  \LevelCPanel{1}{0}{LevelCAdapt}{(b) L1}{Within-Task Adaptation}{\LevelCTopHeight}
  \LevelCPanel{0}{1}{LevelCEvolve}{(c) L2}{Self-Evolution}{\LevelCBottomHeight}
  \LevelCPanel{1}{1}{LevelCRecursive}{(d) L3}{Recursive Self-Improvement}{\LevelCBottomHeight}

  \begin{scope}
    \LevelCAgentFrame
    \node[levelC step] (lc0observe) at (22pt,-56pt) {Observe};
    \node[levelC step] (lc0plan) at (.5\LevelCCardWidth,-56pt) {Plan};
    \node[levelC step] (lc0act) at (\dimexpr\LevelCCardWidth-22pt\relax,-56pt) {Act};
    \draw[levelC arrow] (lc0observe.east) -- (lc0plan.west);
    \draw[levelC arrow] (lc0plan.east) -- (lc0act.west);
    \node[levelC note] at (.5\LevelCCardWidth,-75pt) {preset strategy $W_0$};
  \end{scope}

  \begin{scope}[
      levelC mini/.append style={fill=LevelCAdaptFill!35!white},
      xshift=\dimexpr\LevelCCardWidth+\LevelCColumnGap\relax]
    \LevelCAgentFrame
    \node[levelC step] (lc1observe) at (22pt,-56pt) {Observe};
    \node[levelC step] (lc1plan) at (.5\LevelCCardWidth,-56pt) {Plan};
    \node[levelC step] (lc1act) at (\dimexpr\LevelCCardWidth-22pt\relax,-56pt) {Act};
    \draw[levelC arrow] (lc1observe.east) -- (lc1plan.west);
    \draw[levelC arrow] (lc1plan.east) -- (lc1act.west);
    \draw[levelC arrow] (lc1act.north) -- ++(0,10pt) -|
    node[levelC note,pos=.25,above=1pt] {feedback} (lc1plan.north);
    \node[levelC note] at (.5\LevelCCardWidth,-75pt) {revise workflow $W_t$};
  \end{scope}

  \begin{scope}[
      levelC mini/.append style={fill=LevelCEvolveFill!35!white},
      yshift=\dimexpr-\LevelCTopHeight-\LevelCRowGap\relax]
    \node[levelC symbol,fill=LevelCEvolveFill]
    (lc2update) at (.5\LevelCCardWidth,-43pt) {$U_0$};
    \node[levelC mini,minimum width=\LevelCResourceWidth]
    (lc2before) at (30pt,-70pt) {$\mathcal{A}_k$};
    \node[levelC mini,minimum width=\LevelCResourceWidth]
    (lc2after) at (\dimexpr\LevelCCardWidth-30pt\relax,-70pt) {$\mathcal{A}_{k+1}$};
    \draw[levelC arrow] (lc2before.north) |-
    node[levelC note,anchor=east] {$\mathcal{E}_{\leq k}$} (lc2update.west);
    \draw[levelC arrow] (lc2update.east) -| (lc2after.north);
    \node[levelC note] at (.5\LevelCCardWidth,-90pt)
    {experience-driven resource update};
  \end{scope}

  \begin{scope}[
      levelC mini/.append style={fill=LevelCRecursiveFill!35!white},
      xshift=\dimexpr\LevelCCardWidth+\LevelCColumnGap\relax,
      yshift=\dimexpr-\LevelCTopHeight-\LevelCRowGap\relax]
    \node[levelC symbol] (lc3updater) at (30pt,-46pt) {$U_{k-1}$};
    \node[levelC note] (lc3revision) at (.5\LevelCCardWidth,-46pt) {$\Gamma$};
    \node[levelC symbol,fill=LevelCRecursiveFill]
    (lc3successor) at (\dimexpr\LevelCCardWidth-30pt\relax,-46pt) {$U_k$};
    \node[levelC note] (lc3experience) at (.5\LevelCCardWidth,-31pt) {$\mathcal{E}_{\leq k-1}$};
    \draw[levelC arrow] (lc3updater.east) -- (lc3revision.west);
    \draw[levelC arrow] (lc3experience.south) -- (lc3revision.north);
    \draw[levelC arrow] (lc3revision.east) -- (lc3successor.west);

    \node[levelC mini,minimum width=\LevelCResourceWidth]
    (lc3before) at (30pt,-70pt) {$\mathcal{A}_k$};
    \node[levelC mini,minimum width=\LevelCResourceWidth]
    (lc3after) at (\dimexpr\LevelCCardWidth-30pt\relax,-70pt) {$\mathcal{A}_{k+1}$};
    \coordinate (lc3update) at (.5\LevelCCardWidth,-70pt);
    \draw[levelC arrow] (lc3before.east) --
    node[levelC note,below=2pt] {$\mathcal{E}_{\leq k}$} (lc3after.west);
    \draw[levelC arrow,draw=LevelCRecursive!80!black]
    (lc3successor.south) |- (.5\LevelCCardWidth,-58pt) -- (lc3update);
    \node[levelC note] at (.5\LevelCCardWidth,-90pt)
    {reuse of the revised updater};
  \end{scope}
\end{tikzpicture}%
\endgroup
  \vspace{-6mm}
  \SurveyFigureCaption{Fixed-workflow reference and L1--L3 processes}{%
    (a) At L0, $\mathcal{A}_0$ follows the workflow $W_0$ set before task execution, so $W_t = W_0$ at every step $t$.
    (b) At L1, $\mathcal{A}_0$ can revise the active workflow $W_t$ into $W_{t+1}$ in response to task feedback.
    (c) At L2, the fixed updater $U_0$ uses experience $\mathcal{E}_{\leq k}$ to update model or harness resources used in workflow execution, changing the agent state from $\mathcal{A}_k$ to $\mathcal{A}_{k+1}$.
    (d) At L3, $\Gamma$ uses $\mathcal{E}_{\leq k-1}$ to refine $U_{k-1}$ into $U_k$. The revised updater then uses $\mathcal{E}_{\leq k}$ to update the agent state from $\mathcal{A}_k$ to $\mathcal{A}_{k+1}$.}
  \vspace{-2mm}
  \label{fig:level-characteristics}
\end{figure}

At L1, the agent uses new evidence to adapt its execution of the current task. It may change the active workflow, as illustrated in Fig.~\ref{fig:level-characteristics}(b), or revise or reselect a proposed output or action sequence. For changes to the selected workflow steps or their order, we write
\begin{equation}
  W_{t+1} = \Phi(W_t,o_{t+1},f_t),
  \label{eq:l1-strategy-adaptation}
\end{equation}
where $f_t$ is the feedback available after step $t$ and $\Phi$ is the workflow revision function. A fixed $\Phi$ can produce different workflows as observations and feedback change.

At L2, the agent updates a model or harness resource using experience from its own interactions or training attempts. It retains the updated resource and uses it in a later task, trial, or training round. Human demonstrations, annotations, and external evaluations can provide additional supervision. At stage $k$, the available experience $\mathcal{E}_{\leq k}$ includes current observations, outcomes, and feedback, along with records retrieved from earlier stages. The fixed updater $U_0$ shown in Fig.~\ref{fig:level-characteristics}(c) uses the agent state $\mathcal{A}_k$ from Eq.~\eqref{eq:agent-foundations} and this experience to produce the next state as
\begin{equation}
  \mathcal{A}_{k+1} = U_0(\mathcal{A}_k,\mathcal{E}_{\leq k}).
  \label{eq:l2-resource-evolution}
\end{equation}
The updater can use this experience to propose, apply, and evaluate changes to the model or harness. In subsequent execution, the agent uses the models and harness in $\mathcal{A}_{k+1}$ to carry out the active workflow $W_t$. Several methods described as RSI fall under L2 because they reuse revised harnesses or skill libraries in later stages with a fixed updater~\cite{xu2026videoharness,luo2026video,du2026designer}.

At L3, the agent revises its updater using experience, as illustrated in Fig.~\ref{fig:level-characteristics}(d). At stage $k$, it uses the updater $U_k$ produced at stage $k-1$ to update its state and applies the revision procedure $\Gamma$ to obtain the next updater $U_{k+1}$. The state update is
\begin{equation}
  \mathcal{A}_{k+1} = U_k(\mathcal{A}_k,\mathcal{E}_{\leq k}).
  \label{eq:l3-resource-evolution}
\end{equation}
A separate mechanism can carry out the revision, or the updater can help revise itself. The rules for a fixed procedure $\Gamma_0$ and a self-revising procedure $\Gamma_k$ are
\begin{equation}
  U_{k+1} = \Gamma_0(U_k,\mathcal{A}_k,\mathcal{E}_{\leq k}),
\end{equation}
\begin{equation}
  (U_{k+1},\Gamma_{k+1}) = \Gamma_k(U_k,\mathcal{A}_k,\mathcal{E}_{\leq k}).
\end{equation}
At stage $k+1$, $U_{k+1}$ updates the state from Eq.~\eqref{eq:l3-resource-evolution} using the available experience as follows
\begin{equation}
  \mathcal{A}_{k+2} = U_{k+1}(\mathcal{A}_{k+1},\mathcal{E}_{\leq k+1}).
  \label{eq:l3-recursive-update}
\end{equation}
With either a fixed $\Gamma_0$ or an evolving $\Gamma_k$, this update meets the L3 criterion when it uses a modified component of $U_{k+1}$, such as a retrained evaluator whose scores on trial runs guide the policy update. In the second case, applying $\Gamma_{k+1}$ to a later updater extends the recursion to the procedure itself.

\section{L1 Within-Task Adaptation of Visual Agents}
\label{sec:level1}

As introduced in Section~\ref{sec:taxonomy}, within-task adaptation (L1) describes how a \emph{visual agent} responds to new observations and action outcomes during a task. When the evidence leaves the next decision uncertain, the agent may need a better view before deciding how to proceed. A revised plan can in turn change which capabilities the harness invokes, and feedback during execution can reveal an output or action that needs correction. By monitoring the effects of its actions, the agent can judge whether to proceed, recover, or stop. Table~\ref{tab:within-task-overview} summarizes how these mechanisms act on different parts of execution.

\begin{table*}[!t]
  \centering
  \begingroup
  \SurveyTableStyle
  \SurveyTableCaption{Within-Task Adaptation Mechanisms in Visual Agents.}{tab:within-task-overview}
      \vspace{-2mm}
  \setlength{\tabcolsep}{.5pt}
  \newcommand{\LoneOverviewContents}{%
    \SurveyTableTopRule
    \SurveyTableRowColor{SurveyTableHeader}
    \textbf{Dimension} & \textbf{Mechanism} & \textbf{Target} &
    \textbf{Key Operation} & \textbf{Representative Works} \\
    \SurveyTableHeaderRule
    \SurveyTableRowColor{white}
    & Adaptive Observation
    & Visual Input
    & Crop / Zoom / Select Viewpoints
    & \cite{AAAI2026Guieyes,CVPR2026ActiveVLA,ACL2026DocVStar} \\
    \SurveyTableRowColor{SurveyTableStripe}
    & Re-Grounding
    & Visual Correspondence
    & Refine Entity / Region Matches
    & \cite{added_li2026deepscan,CVPR2026DRSGUI,bundele2026trackers} \\
    \SurveyTableRowColor{white}
    \multirow{-3}{=}{\raggedright\bfseries Adaptive Perception}
    & Re-Observation
    & Prior Observations
    & Revisit Intervals / Views
    & \cite{CVPR2026LensWalk,ICLR2026Videozoomer,NeurIPS2026VideoTrees} \\
    \midrule[.3pt]
    \SurveyTableRowColor{SurveyTableStripe}
    & Goal and Subgoal Revision
    & Intermediate Objectives
    & Revise / Add Subgoals
    & \cite{added_huang2023innermonologue,xie2026preact,NeurIPS2024GenArtist} \\
    \SurveyTableRowColor{white}
    & Local Plan Repair
    & Local Plan Segments
    & Patch Affected Segments
    & \cite{zhong2026actionengine,NeurIPS2026VLACorrector,AAAI2026TCoT} \\
    \SurveyTableRowColor{SurveyTableStripe}
    & Replanning
    & Remaining Plan Horizon
    & Recompute Remaining Plan
    & \cite{ICLR2026Sparse,liu2026pilot,ACL2026ModelBasedImagination} \\
    \SurveyTableRowColor{white}
    \multirow{-4}{=}{\raggedright\bfseries Feedback-Driven Planning}
    & Constraint and Feasibility Reassessment
    & Candidates / Planning Assumptions
    & Check Feasibility / Reassess Assumptions
    & \cite{mishra2026tampagent,NeurIPS2026APIVOT,ICLR2026Simulation} \\
    \midrule[.3pt]
    \SurveyTableRowColor{SurveyTableStripe}
    & Model and Agent Routing
    & Models / Agents
    & Select or Configure Models / Route Agents
    & \cite{ICLR2026MATA,AAAI2026Genmac,NeurIPS2026VLAPro} \\
    \SurveyTableRowColor{white}
    & Tool and Skill Orchestration
    & Tool Calls / Skills
    & Select Tools / Switch Skills
    & \cite{ICLR2026Adareasoner,NeurIPS2026MMSkills,ACL2026Breaking} \\
    \SurveyTableRowColor{SurveyTableStripe}
    \multirow{-3}{=}{\raggedright\bfseries Dynamic Harness Orchestration}
    & Multi-Agent Coordination
    & Agent Interactions
    & Review / Hand Off / Resolve Disputes
    & \cite{added_yu2026mact,lassoued2026orca,CVPR2026A} \\
    \midrule[.3pt]
    & Textual and Structured Output Correction
    & Answers / Predictions
    & Refine Responses / Structured Outputs
    & \cite{jeong2026draft,CVPR2026AgentDet,zhang2026pixelcraft} \\
    \SurveyTableRowColor{SurveyTableStripe}
    & Visual Artifact Revision
    & Artifacts / Prompts / Code
    & Edit Prompts / Content / Code
    & \cite{ICLR2026Visualprompter,CVPR2026MMReCoder,ICLR2026Webgenagent} \\
    \SurveyTableRowColor{white}
    \multirow{-3}{=}{\raggedright\bfseries Output and Action Refinement}
    & Action Correction
    & Proposed Actions / Chunks
    & Revise / Reselect Actions
    & \cite{CVPR2026Diagnose,added_kim2026race} \\
    \midrule[.3pt]
    \SurveyTableRowColor{SurveyTableStripe}
    & State Tracking
    & State and Progress Records
    & Update Records / Track Progress
    & \cite{ICLR2026Planning,CVPR2026MemoryAugmented,liu2026memory} \\
    \SurveyTableRowColor{white}
    & Monitoring
    & Execution Conditions
    & Check Prerequisites / Effects
    & \cite{ACL2026VeriGUI,guo2026qreflectorlearningguiaction,yang2026task} \\
    \SurveyTableRowColor{SurveyTableStripe}
    & Recovery
    & Task-Valid State
    & Compensate / Roll Back / Reset
    & \cite{ICLR2026LongHorizonUI,CVPR2026FLARE,CVPR2026REACH,zhang2026imagining} \\
    \SurveyTableRowColor{white}
    \multirow{-4}{=}{\raggedright\bfseries Execution Monitoring and Control}
    & Termination
    & Continuation Decision
    & Accept / Continue / Stop
    & \cite{CVPR2026ProofofPerception,AAAI2026Talk2Image} \\
    \SurveyTableBottomRule
  }
  \ifdefined\LoneOverviewBox\else\newsavebox{\LoneOverviewBox}\fi
  \sbox{\LoneOverviewBox}{%
    \begin{tabular}{p{.14\linewidth}lllc}\LoneOverviewContents\end{tabular}}
  \typeout{Within-task table: \the\wd\LoneOverviewBox; text: \the\linewidth}
  \ifdim\wd\LoneOverviewBox>\linewidth
    \PackageError{LoneOverviewTable}{Single-Line Table Exceeds Text Width}%
    {Shorten A Label Without Scaling The Font.}
  \fi
  \setlength{\tabcolsep}{\dimexpr\tabcolsep+(\linewidth-\wd\LoneOverviewBox)/10\relax}
  \begin{tabular}{p{.14\linewidth}lllc}
    \LoneOverviewContents
  \end{tabular}

  \SurveyTableDescription{%
    The five dimensions correspond to the L1 categories in Section~\ref{sec:level1}.
  }
  \endgroup
  \vspace{-6mm}
\end{table*}

\subsection{Adaptive Perception}
\label{sec:l1-perception}

\begin{figure}[!t]
    \centering
    \includegraphics[width=\columnwidth]{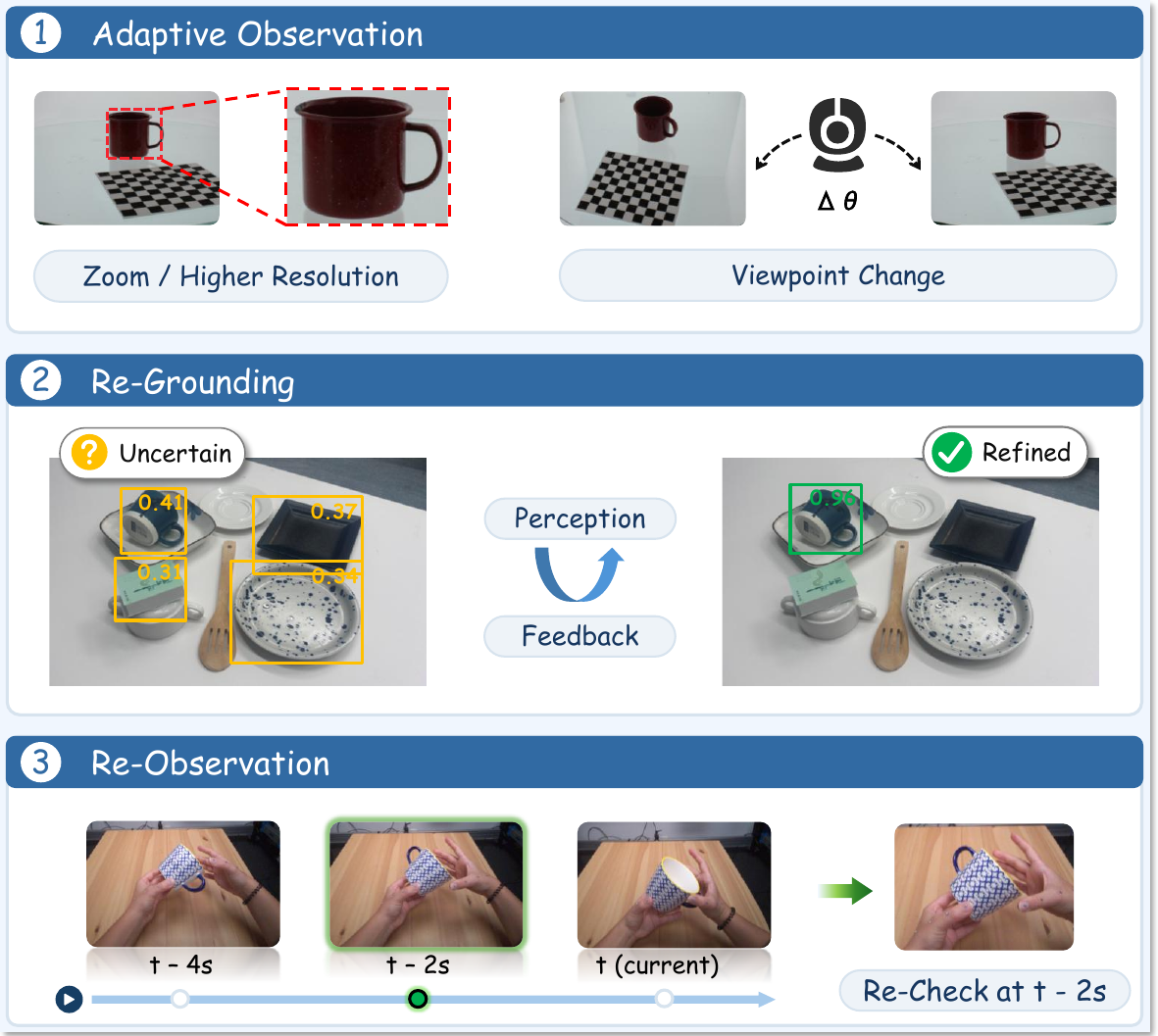}
    \vspace{-6mm}
    \SurveyFigureCaption{Adaptive perception through evidence refinement}{Adaptive observation (1) obtains additional evidence from a higher-resolution crop or a new viewpoint, with \(\Delta\theta\) denoting the change in viewing angle. Re-grounding (2) uses feedback to refine an uncertain target match, as illustrated by the higher match score for the red mug. Re-observation (3) revisits the frame at \(t-2\,\mathrm{s}\) to recheck the mug.}
    \vspace{-2mm}
    \label{fig:adaptive-perception}
\end{figure}

An observation that supports one decision may leave the next unresolved. A more informative view can supply missing evidence, and feedback can help the agent check whether the evidence points to the intended target. If later observations raise new questions, the agent can return to earlier evidence for another look. Fig.~\ref{fig:adaptive-perception} illustrates these adjustments.

\noindent$\bullet$\ \textbf{Adaptive Observation.} To obtain the evidence needed for the current decision, the agent can adjust what it inspects or how closely it looks. Doc-V$^*$~\cite{ACL2026DocVStar} actively fetches cached high-resolution page tokens after inspecting thumbnails. Look Less, Reason More~\cite{ACL2026LookLess} learns when to invoke selective pixel operations. Reason with Thumbnails, Answer with Focus~\cite{ICML2026Reason} reasons over thumbnails before requesting focused crops and accesses higher-resolution evidence when thumbnails are insufficient. GUI-Eyes~\cite{AAAI2026Guieyes} selects targeted crops before grounding and reasons over the enlarged views. ActiveVLA~\cite{CVPR2026ActiveVLA} selects task-critical virtual views from a point cloud. SaPaVe~\cite{CVPR2026SaPaVe} instead controls a physical camera to expose useful evidence for manipulation.

\noindent$\bullet$\ \textbf{Re-Grounding.} A clear view can still leave the target uncertain. The agent can use feedback to check which object or region corresponds to the target or to refine its location. DeepScan~\cite{added_li2026deepscan} refines localized evidence views through feedback between a vision-language model and visual experts. DRS-GUI~\cite{CVPR2026DRSGUI} searches alternative interface regions and revises its focus using region-quality feedback. Visual Self-Refine~\cite{ICLR2026VisualSelfRefine} corrects chart-point anchors rendered on the image before decoding the corresponding values. MOSEv2~\cite{ding2025mosev2} highlights the challenges of severe occlusion and intermittent target visibility in video object segmentation. Dual-Agent~\cite{bundele2026trackers} uses temporal context to verify target identity and selectively re-localizes the target after tracking drift or a missed reappearance.

\noindent$\bullet$\ \textbf{Re-Observation.} When a later observation leaves a question unresolved, the agent may revisit earlier evidence or inspect a familiar scene from another viewpoint. VideoTreeSearch~\cite{NeurIPS2026VideoTrees} uses \texttt{zoom\_out} and \texttt{shift} to revisit earlier temporal choices. LensWalk~\cite{CVPR2026LensWalk} links observations to timestamps and entities and resamples ambiguous intervals more densely. VideoZoomer~\cite{ICLR2026Videozoomer} uses temporal zoom to revisit selected intervals at a higher frame rate after a sparse overview. When retrieved visual memories provide insufficient evidence, RAVEN~\cite{hu2026raven} issues another semantic, spatial, or temporal query. ForkCast~\cite{wen2026forkcast} queries a frozen policy with candidate camera views and scores the resulting action continuations to decide whether to re-observe from a new viewpoint.

\subsection{Feedback-Driven Planning}
\label{sec:l1-planning}

New visual evidence can change which subgoal is worth pursuing. When the objective still holds, execution feedback can reveal a step that needs repair or show that the remaining plan needs reconsideration. Observed and predicted outcomes also help the agent judge whether candidate actions are feasible and whether its constraints need revision.

\noindent$\bullet$\ \textbf{Goal and Subgoal Revision.} The overall goal can remain the same as the agent replaces or adds subgoals in light of new evidence. Inner Monologue~\cite{added_huang2023innermonologue} incorporates updated object detections and success feedback into its language planner to revise manipulation subgoals as initially occluded objects become visible. PreAct-Nav~\cite{xie2026preact} compares real observations with expected visual cues and replaces the active navigation subgoal when execution feedback indicates completion or contradicts the current objective. For image generation and editing, GenArtist~\cite{NeurIPS2024GenArtist} inspects generated images and adds corrective editing subgoals to its planning tree to address missing objects, incorrect attributes, or spatial errors.

\noindent$\bullet$\ \textbf{Local Plan Repair.} Observed or anticipated difficulties with part of a plan can prompt the agent to revise the affected steps and continue the task. ActionEngine~\cite{zhong2026actionengine} includes a Patcher that repairs faulty code or re-grounds failed GUI actions within the current plan separately from memory updates retained for future tasks, with the Patcher disabled in the main evaluation. VLA-Corrector~\cite{NeurIPS2026VLACorrector} replaces stale remaining actions with a replanned chunk after persistent visual-dynamics disagreement. TCoT~\cite{AAAI2026TCoT} checks task completion after a prescribed period of fixed global guidance and, if the task remains unfinished, generates local gripper trajectories and actions for recovery. SeqWalker~\cite{AAAI2026Seqwalker} checks observed subinstruction progression against the instruction order, backtracking and trying an alternative action when the check fails. ForkCast~\cite{wen2026forkcast} compares repaired and nominal arm trajectories at the same prediction horizon and adopts the repair if it yields more progress.

\noindent$\bullet$\ \textbf{Replanning.} If new observations or predicted outcomes favor a different course of action, the agent can revise its remaining steps. In visual model-predictive control, Sparse Imagination~\cite{ICLR2026Sparse} uses sparse visual tokens to replan from fresh observations. PILOT~\cite{liu2026pilot} combines video-derived visual subgoals, policy priors, and a world model to refine action sequences through receding-horizon planning. IMPLEMENT~\cite{ACL2026ModelBasedImagination} refines decisions by comparing predicted object-centric futures. R-WoM~\cite{ICLR2026RWoM} grounds candidate computer-use simulations in retrieved tutorials to inform planning. V-ABS~\cite{ICML2026VABS} revises beam-search preferences using executed visual-tool outcomes and entropy-weighted observer feedback. For image generation, Talk2Image~\cite{AAAI2026Talk2Image} uses evaluation feedback to revise its intent representation and task graph, then reschedules affected generation or editing branches for re-execution.

\noindent$\bullet$\ \textbf{Constraint and Feasibility Reassessment.} Observations and predictions help test candidate actions against task constraints. A failed check may warrant a different action or a revision to the planner's assumptions. VLMFP~\cite{ICLR2026Simulation} revises PDDL action rules when symbolic execution disagrees with consequences predicted by a visual simulator. TAMPAgent~\cite{mishra2026tampagent} couples visual belief updates with grasp, placement, and motion checks whose failures can trigger alternative candidates, replanning, or a revised scene hypothesis. APIVOT~\cite{NeurIPS2026APIVOT} checks spatial constraints and remaining free space in imagined visual states while constructing a candidate plan.

\subsection{Dynamic Harness Orchestration}
\label{sec:l1-orchestration}

Deciding what to do next also raises the question of which capabilities the harness should invoke. Evidence from one operation may reveal the need for a specialist or a different tool or skill. When agents contribute separate results, further review can help them resolve disagreements or identify work to return for correction.

\noindent$\bullet$\ \textbf{Model and Agent Routing.} Which model or agent is suited to the next operation can change as observations and intermediate results arrive. MATA~\cite{ICLR2026MATA} selects the next visual reasoning agent using tool outputs and verification feedback in shared memory and reroutes after agent failures. GenMAC~\cite{AAAI2026Genmac} selects a specialized correction agent based on consistency, temporal, or spatial errors detected in generated videos. VLA-Pro~\cite{NeurIPS2026VLAPro} uses current visual observations and execution history to retrieve and fuse task adapters per action chunk, restoring the base model after each chunk.

\noindent$\bullet$\ \textbf{Tool and Skill Orchestration.} The agent can use the evidence available at each step to select and combine tools or skills. When feedback shows that a skill no longer fits the task state, it can switch skills or return to direct action. AdaReasoner~\cite{ICLR2026Adareasoner} conditions successive tool selections on intermediate observations, composing perception, image manipulation, and calculation tools within a visual reasoning trajectory. Earth-Agent~\cite{ICLR2026Earthagent} incorporates visual, spectral, and quantitative tool outputs into its interaction history to guide subsequent tool choices and arguments for Earth observation tasks. MMSkills~\cite{NeurIPS2026MMSkills} allows the agent to act directly or consult a multimodal skill in a temporary reasoning branch, which aligns reference evidence with the live state and returns procedural guidance. KnowAct-GUIClaw~\cite{li2026knowact} checks whether retrieved GUI skills apply and lead to expected states, with direct GUI actions as a fallback when state checks fail. AgentVLN~\cite{xin2026agentvln} selects perception or navigation skills from current visual observations and history, temporarily using fine-grained corrective actions when no suitable visual waypoint is available. SkillNav~\cite{ACL2026Breaking} uses current visual observations and past actions to identify the active navigation subgoal and select its specialized skill policy.

\noindent$\bullet$\ \textbf{Multi-Agent Coordination.} When agents disagree on a result or find an error, they may need to review each other's work to reconcile their findings or identify a step to return for correction. MACT~\cite{added_yu2026mact} uses a judgment agent to return erroneous planning or execution steps to the responsible agent for correction before forwarding verified results to the answer agent. ORCA~\cite{lassoued2026orca} triggers stress testing when thinker and specialist answers disagree, escalating to thesis--antithesis debate and adjudication if the specialist fails the checks. A4VL~\cite{CVPR2026A} uses cross-review to narrow the team and start another exploration round when agents disagree on a video question.

\subsection{Output and Action Refinement}
\label{sec:l1-correction}

Visual evidence and evaluation feedback can reveal where an initial output falls short of the task requirements and guide its revision. Likewise, changes in the observed state can lead the agent to revise an upcoming action.

\noindent$\bullet$\ \textbf{Textual and Structured Output Correction.} Comparing an initial answer or a structured prediction with visual evidence can reveal errors. Given an initial answer, Draft and Refine~\cite{jeong2026draft} obtains visual-expert evidence, renders it onto the image, and re-queries the model to select a response with greater measured visual dependence. PixelCraft~\cite{zhang2026pixelcraft} reviews an initial answer's reasoning and tool trace, then uses identified errors to guide a second answering attempt. Visual Self-Refine~\cite{ICLR2026VisualSelfRefine} iteratively corrects pixel anchors on charts before extracting the corresponding values. For structured detections, AgentDet~\cite{CVPR2026AgentDet} uses its judge role to refine proposed object boxes. Dual-Agent~\cite{bundele2026trackers} corrects the target visibility flag and bounding box after its verifier rejects the current tracking hypothesis, then re-prompts the tracker with accepted corrections.

\noindent$\bullet$\ \textbf{Visual Artifact Revision.} Inspecting a generated artifact may reveal a defect the agent can correct with a direct edit. Feedback can also guide revisions to the prompts or code that produced it. Prompt-based methods revise prompts to address missing image concepts or incorporate visual, audio, and contextual critiques of generated videos~\cite{ICLR2026Visualprompter,CVPR2026VISTA}. Direct artifact correction includes inpainting for localized distortions~\cite{added_shen2026agenticretoucher}, pose refinement using visual correspondence and physical-plausibility checks~\cite{ICLR2026Scenethesis}, simulated-audience-guided revisions to film editing and audiovisual rhythm~\cite{ICLR2026FilMaster}, and successive image edits guided by visual verification~\cite{ICLR2026Generative}. For chart code, MM-ReCoder~\cite{CVPR2026MMReCoder} learns to correct code over multiple turns using execution feedback. CoDA~\cite{ICLR2026CoDA} guides revisions through debugging and self-reflection. WebGen-Agent~\cite{ICLR2026Webgenagent} uses screenshot critiques and GUI tests to refine webpage code and select the best code state.

\noindent$\bullet$\ \textbf{Action Correction.} A failed action or a change in the observed state can prompt the agent to adjust an upcoming movement or select a different sequence. ViFailback~\cite{CVPR2026Diagnose} transforms manipulation failure diagnoses into textual and visual-symbolic guidance for corrective movements. After asynchronous inference, RACE~\cite{added_kim2026race} uses the robot's measured position and velocity to select an action chunk the robot can follow smoothly, then retimes it under kinodynamic constraints.

\subsection{Execution Monitoring and Control}
\label{sec:l1-state}

Tracking changes in the environment and task progress helps the agent assess proposed actions and check their effects. These checks can reveal when recovery is needed, and evidence about the current result helps the agent decide whether to continue or end the attempt within the remaining budget.

\noindent$\bullet$\ \textbf{State Tracking.} New observations and action outcomes keep the agent's record of the environment and task progress up to date. Embodied Perception Memory (EPM)~\cite{ICLR2026Planning} updates a textual environment representation from egocentric images, allowing the planner to account for changes in object locations and states. OctMem-Agent~\cite{CVPR2026MemoryAugmented} updates a 3D octree map with successive RGB-D observations, recording explored regions and unexplored frontiers for subsequent aerial navigation decisions. ATMem~\cite{liu2026memory} updates task records from GUI observations and operation outcomes, marking data items as pending, completed, or skipped to guide subsequent actions.

\noindent$\bullet$\ \textbf{Monitoring.} Before acting, the agent can check whether a proposed action satisfies the conditions for execution. Afterward, observations help it verify the intended effects and check whether task constraints still hold. At deployment, the trained OS-Oracle critic~\cite{CVPR2026OSOracle} checks proposed actions and allows bounded regeneration. MaDS~\cite{ACL2026MaDS} compares competing assessments of a proposed action before execution. TaskVM~\cite{yang2026task} separates execution from verification of desired outcomes, preservation constraints, and observed state. VeriGUI~\cite{ACL2026VeriGUI} compares expected effects with new screenshots and uses detected failures for corrective reasoning. Q-Reflector~\cite{guo2026qreflectorlearningguiaction} checks generated action pre- and post-conditions against paired screenshots, distinguishing execution success from task validity. Self-Grounded Verification~\cite{ICLR2026Lets} evaluates behavior against priors elicited without inspecting the candidate trajectory and uses these judgments to guide refinement. OS-Sentinel~\cite{ACL2026OSSentinel} combines formal trace checks with a contextual visual judge and evaluates detection primarily on frozen trajectories, with a comparison to live execution.

\noindent$\bullet$\ \textbf{Recovery.} A failed action can leave the environment in a state that prevents the next step. Recovery restores the conditions the agent needs to resume the task. LongHorizonUI~\cite{ICLR2026LongHorizonUI} attempts local compensation before rolling back to a committed snapshot. TaskVM~\cite{yang2026task} distinguishes reversible, compensatable, and irreversible effects when supporting user-governed rollback. REACH~\cite{CVPR2026REACH} combines rollback with negative action prompts under assumptions of reversible transitions and sufficiently early detection. FLARE~\cite{CVPR2026FLARE} switches from task execution to a learned reset skill after a state-breaking failure, restores task-valid conditions, and then returns control to the task policy. CoRe~\cite{zhang2026imagining} uses counterfactual visual rollouts after an execution deviation to select which displaced objects need restoration, then restores the selected objects and realigns the robot before resuming the frozen task policy.

\noindent$\bullet$\ \textbf{Termination.} The agent can end an attempt once the evidence is sufficient to accept the result. If another step could resolve uncertainty or improve the result, the remaining budget determines whether the agent can continue. A4VL~\cite{CVPR2026A} uses cross-review and relevance ranking to accept consensus or initiate further exploration. MSSR~\cite{ICLR2026Pursuing} removes redundant spatial evidence and requests missing information until an adequate set supports the task. Proof-of-Perception~\cite{CVPR2026ProofofPerception} uses calibrated conformal uncertainty to decide whether to stop or gather more evidence within a compute budget, with execution certificates guiding any further evidence gathering. Draft and Refine~\cite{jeong2026draft} uses measured visual utilization as a grounding-based stopping proxy and retains the draft when no visual expert improves that measure. Talk2Image~\cite{AAAI2026Talk2Image} accepts the result when evaluation feedback is stable and sufficiently high or stops refining once it exhausts its retry allowance.

\section{L2 Self-Evolving Visual Agents}
\label{sec:level2}

In L2 self-evolution, experience from the agent's own attempts guides updates to resources that it keeps and uses in later tasks, trials, or training rounds. Training can carry what the agent has learned into the models it uses later, and retained memories and procedures can change how its harness handles subsequent tasks. The outcomes of those tasks may in turn guide revisions to the harness or reveal what the agent should learn from next. Table~\ref{tab:self-evolution-overview} relates each mechanism to the experience it uses and the resources it updates.

\begin{table*}[!t]
  \centering
  \begingroup
  \SurveyTableStyle
  \SurveyTableCaption{Experience-Driven Updates in Self-Evolving Visual Agents.}{tab:self-evolution-overview}
  \vspace{-2mm}
  \setlength{\tabcolsep}{.5pt}
  \hyphenpenalty=10000\exhyphenpenalty=10000

  \newcommand{\LtwoOverviewContents}{%
    \SurveyTableTopRule
    \SurveyTableRowColor{SurveyTableHeader}
    \textbf{Experience Source} & \textbf{Update Process} & \textbf{Retained Resource} & \textbf{Update Target} & \textbf{Subsequent Use} & \textbf{Representative Works} \\
    \SurveyTableHeaderRule
    \SurveyTableRowColor{white}
    Outputs / Feedback
    & Supervised / Reinforcement Learning
    & Model Parameters
    & Model
    & Visual Reasoning / Generation
    & \cite{AAAI2026HistoryAware,ACL2026ReLook,ICLR2026DeepEyes,tan2026getting,CVPR2026GUISAGE} \\
    \SurveyTableRowColor{SurveyTableStripe}
    Own Rollouts / Outcomes
    & Policy Learning
    & Policy Parameters
    & Model
    & Action / Tool Selection
    & \cite{ICLR2026ComputerRL,CVPR2026Learningto,NeurIPS2026ToolCUA} \\
    \SurveyTableRowColor{white}
    Interaction Rollouts
    & World-Model Adaptation
    & World-Model Parameters
    & Model
    & Later Planning / Control
    & \cite{AAAI2026Worldagen,levy2026simulation,zhou2026benchmarking} \\
    \SurveyTableRowColor{SurveyTableStripe}
    Procedural Guidance
    & Behavior Distillation
    & Policy Parameters
    & Model
    & Action Generation
    & \cite{liu2026skilllens,fan2026skill} \\
    \SurveyTableRowColor{white}
    Successful Trajectories
    & Experience Encoding
    & Experience Records
    & Memory
    & Execution Guidance
    & \cite{CVPR2026Dejavu,CVPR2026Experience,zhu2026hybrid} \\
    \SurveyTableRowColor{SurveyTableStripe}
    Errors / Critiques
    & Guideline Induction
    & Diagnostic Guidance
    & Memory
    & Decision Guidance
    & \cite{CVPR2026ViLoMem,ACL2026RSMeM,xie2026mineevolve,jiang2026manimagent} \\
    \SurveyTableRowColor{white}
    User Interactions
    & Knowledge Encoding
    & User Knowledge
    & Memory
    & Personalized Guidance
    & \cite{AAAI2026PerTouch,ICLR2026Embodied} \\
    \SurveyTableRowColor{SurveyTableStripe}
    Feedback on Memory Use
    & Policy Revision / Validation
    & Retrieval Policy
    & Memory
    & Evidence Selection
    & \cite{tian2026abot,che2026lvmh} \\
    \SurveyTableRowColor{white}
    Successful Executions
    & Procedure Extraction
    & Callable Procedures
    & Skills / Tools
    & Procedure Reuse
    & \cite{zhao2026echopath,ICLR2026Transductive} \\
    \SurveyTableRowColor{SurveyTableStripe}
    Skill Failures
    & Reflective Revision
    & Skill Guidance
    & Skills / Tools
    & Later Execution
    & \cite{ju2026embodiskill,chen2026reflect,jiang2026visualskill} \\
    \SurveyTableRowColor{white}
    Tool-Use Traces
    & Tool Abstraction
    & Callable Tools
    & Skills / Tools
    & Tool Reuse
    & \cite{CVPR2026META,fan2026evolving} \\
    \SurveyTableRowColor{SurveyTableStripe}
    Execution Feedback
    & Prompt / Input Revision
    & Prompts / Input Code
    & Orchestration
    & Later Model Inputs
    & \cite{luo2026evogen,feng2026enactopt,xu2026videoharness} \\
    \SurveyTableRowColor{white}
    Tool / Task Outcomes
    & Workflow / Control Updates
    & Graphs / Code / Scores
    & Orchestration
    & Execution Control
    & \cite{yang2026gui,zhou2026automating,wang2026vevo,chen2026kpharness,jiang2026harnessmmv} \\
    \SurveyTableRowColor{SurveyTableStripe}
    Rollout Signals
    & Data Acquisition / Refinement
    & Training Examples
    & Data / Tasks
    & Later Training
    & \cite{AAAI2026CorrectAD,CVPR2026HATS,ICLR2026Selfimprovingb,karli2026recall} \\
    \SurveyTableRowColor{white}
    Task Outcomes
    & Task / Curriculum Updates
    & Tasks / Curricula
    & Data / Tasks
    & Later Practice
    & \cite{added_sun2026seagent,NeurIPS2026Autonomous,fang2026find,li2026sees,kang2026simworld} \\
    \SurveyTableBottomRule
  }
  \newsavebox{\LtwoOverviewBox}
  \sbox{\LtwoOverviewBox}{\begin{tabular}{lllllc}\LtwoOverviewContents\end{tabular}}
  \typeout{L2 overview table: \the\wd\LtwoOverviewBox; text: \the\linewidth}
  \ifdim\wd\LtwoOverviewBox>\linewidth
    \PackageError{LtwoOverviewTable}{Single-line overview table exceeds text width}%
    {Shorten a label without scaling the font.}
  \fi
  \begingroup
  \setlength{\tabcolsep}{\dimexpr\tabcolsep+(\linewidth-\wd\LtwoOverviewBox)/12\relax}
  \begin{tabular}{lllllc}\LtwoOverviewContents\end{tabular}
  \endgroup

  \SurveyTableDescription{%
    \textit{Retained Resource} identifies the specific resource kept after the update, and \textit{Update Target} gives the corresponding L2 category.
    The retained resource is used in a later task, a separate attempt at the same task, or a subsequent training round.
  }
  \endgroup
    \vspace{-6mm}
\end{table*}

\subsection{Model-Level Evolution}
\label{sec:l2-models}

Feedback on an agent's outputs or trajectories can supply training signals for the models it uses in later attempts, with demonstrations or labels providing additional supervision. As these updates change how the agent reasons or acts, evaluations can reveal whether it retains earlier capabilities. Experience can also refine the predictions that a world model supplies for planning or policy learning.

\noindent$\bullet$\ \textbf{Model Cognition Updates.} Training with reasoning and generation feedback can change how the model later interprets evidence or produces outputs. History-Aware Reasoning~\cite{AAAI2026HistoryAware} converts incorrect episode predictions into teacher-generated correction guidelines for subsequent reasoning training. On a fixed training set, GUI-SAGE~\cite{CVPR2026GUISAGE} samples reasoning and action candidates from its current policy, adds a self-explanation conditioned on a labeled GUI action, and scores the candidates to update the policy. Counterfactual VLA~\cite{CVPR2026Counterfactual} selects informative rollouts and uses expert actions to obtain teacher-labeled reasoning corrections. ReLook~\cite{ACL2026ReLook} trains a code-generation model with rendered web-code outcomes and multimodal critique before deploying it with fixed weights. DeepEyes~\cite{ICLR2026DeepEyes} optimizes interleaved visual reasoning and inspection trajectories through reinforcement learning, retaining the acquired behavior in its model. SpatialCritic~\cite{tan2026getting} trains a tool-using evaluator on separately constructed spatial preference data and uses it as a fixed reward model to update image generators through reinforcement learning.

\noindent$\bullet$\ \textbf{Action and Control Model Updates.} Outcomes from the agent's own actions can guide policy updates that affect which actions or tools it chooses on later attempts. EvoCUA~\cite{NeurIPS2026EvoCUA} uses verified rollouts and diagnosed failures to supervise later controller training. RePO-VLA~\cite{NeurIPS2026RePOVLA} uses recovery episodes annotated with progress and value labels for policy training. ComputerRL~\cite{ICLR2026ComputerRL} alternates reinforcement and supervised updates when training its GUI policy. CGL~\cite{CVPR2026CGL} combines supervised and reinforcement updates with ground-truth action rewards. VisTA~\cite{CVPR2026Learningto} trains a model to select visual tools based on task outcomes. ToolCUA~\cite{NeurIPS2026ToolCUA} combines supervised and reinforcement learning to train a model that selects between GUI actions and structured tool calls.

\noindent$\bullet$\ \textbf{Capability Retention.} Learning a new task can weaken performance on earlier ones. Comparing results across tasks before and after an update helps establish which capabilities persist and what the model has learned. Liu et al.~\cite{added_liu2026continualvla} identify benefits from small replay buffers in sequential VLA manipulation learning while distinguishing performance loss from knowledge recoverable through later fine-tuning. Zhu et al.~\cite{zhu2026can} observe severe forgetting under naive sequential fine-tuning across ten physical-robot tasks and substantially better retention with configured experience replay. CGL~\cite{CVPR2026CGL} addresses continual GUI learning through entropy-guided balancing of supervised and reinforcement updates and gradient-conflict control. Dream to Remember~\cite{tharwat2026dream} trains a steering module around a frozen generalist robot policy and rehearses earlier tasks through a frozen world model, retaining changes in the steering module. Raghavan et al.~\cite{raghavan2026does} use perturbed instructions to show that retained robotic task success need not preserve reliable language grounding.

\noindent$\bullet$\ \textbf{World-Model Evolution.} Observations from exploration and task execution can guide updates to world models used in later planning or control. WorldAgen~\cite{AAAI2026Worldagen} uses exploratory transitions to adapt the shared world-model backbone for later trials in the same scene while keeping its action-prediction head fixed. SimDist~\cite{levy2026simulation} updates a pretrained latent dynamics model from on-policy robot rollouts between batches of real-world trials and uses the revised model for subsequent planning. Zhou et al.~\cite{zhou2026benchmarking} evaluate continual world-model learning by updating a shared dynamics backbone through online interaction and reusing it across successive simulated manipulation tasks.

\subsection{Memory-Driven Capability Evolution}
\label{sec:l2-memory}

In memory-driven evolution, visual agents use persistent memory to carry experience across tasks and trials. To build this memory, they can retain selected trajectories and distill guidance by reflecting on successes and failures. As experience accumulates, screening and consolidation shape which records remain available and how they are organized. On subsequent attempts, agents can use the current goal and visual observations to retrieve relevant records and apply them during execution.

\noindent$\bullet$\ \textbf{Experience Acquisition and Reflection.} What the agent records from an interaction shapes the evidence it can use later. Reflection can turn that evidence into guidance for execution. Dejavu~\cite{CVPR2026Dejavu} retains visual keys and actions from rollouts whose final states meet a goal-similarity threshold. Echo~\cite{CVPR2026Experience} preserves successful trajectories with provenance identifying their source environment and model version. AtlasVA~\cite{wang2026atlasva} uses privileged simulator state to construct its atlas and updates spatial heatmaps, visual exemplars, and symbolic skills from trajectory statistics to guide later actions and shape rewards. ViLoMem~\cite{CVPR2026ViLoMem} uses ground-truth-assisted error attribution to develop separate visual and logical diagnostic guidelines. RSMeM~\cite{ACL2026RSMeM} stores constraints distilled from failed tool-use traces in memory and retrieves them for later attempts at the same instance. DigimonGPT~\cite{AAAI2026DigimonGPT} retains procedural memories from successful tool workflows. Agent-sama~\cite{AAAI2026Agentsama} stores interface-state graphs and action records from completed interactions for later tasks. PerTouch~\cite{AAAI2026PerTouch} retains confirmed editing parameters alongside scene semantics. Kwon et al.~\cite{ICLR2026Embodied} retain personalized object meanings and routines in the agent's memory. OctoT2I~\cite{CVPR2026OctoT2I} builds a persistent knowledge base through exploratory image generation and scoring and reuses the resulting tool-capability profiles and exploration records to select generators for later requests.

\noindent$\bullet$\ \textbf{Memory Management.} A new record can reinforce an earlier one or bring it into question. Screening and consolidation determine which experience remains available for reuse as the store grows. ViLoMem~\cite{CVPR2026ViLoMem} merges related diagnostic guidelines or creates new schemas. MineEvolve~\cite{xie2026mineevolve} evaluates candidate procedures and remedies for completeness, applicability, executability, specificity, and conflicts before retaining them. HyMEM~\cite{zhu2026hybrid} consolidates successful trajectories into persistent strategy and attribute graphs while keeping refreshed task-local guidance separate. Conflict-Aware Memory~\cite{ACL2026ConflictAware} adds successful robot experiences to its knowledge base and prunes retrieval links that may lead to conflicts. ECHO~\cite{NeurIPS2026ECHO} commits validated trajectories to a long-term hierarchy and restructures it through idle-time clustering and consolidation. MAGNET~\cite{ACL2026MAGNET} retains new interface appearances and successful workflow abstractions, with access governed by frequency and recency. ManimAgent~\cite{jiang2026manimagent} stores explanations of successes and failures alongside rendered keyframes and uses scores from a fixed VLM to select records for reuse across animation tasks.

\noindent$\bullet$\ \textbf{Memory Retrieval and Reuse.} The current goal and visual evidence help the agent retrieve applicable records and reuse them to guide execution. ViLoMem~\cite{CVPR2026ViLoMem} retrieves acquired guidelines through image matching and question-conditioned reranking. HyMEM~\cite{zhu2026hybrid} expands visually and textually matched nodes through its experience graph. DigimonGPT~\cite{AAAI2026DigimonGPT} adapts replay of retained procedural experience to question complexity. PerTouch~\cite{AAAI2026PerTouch} retrieves scene-conditioned editing preferences for a new image. In an eight-split OpenEQA run, ABot-AgentOS~\cite{tian2026abot} proposes candidate rules from memory failures and checks for gains and regressions, retaining a single retrieval rule that prioritizes observed object records by room and last-seen evidence for reuse in later splits.

\subsection{Skill and Tool-Level Evolution}
\label{sec:l2-skills}

\begin{figure}[!t]
  \centering
  \includegraphics[width=\columnwidth]{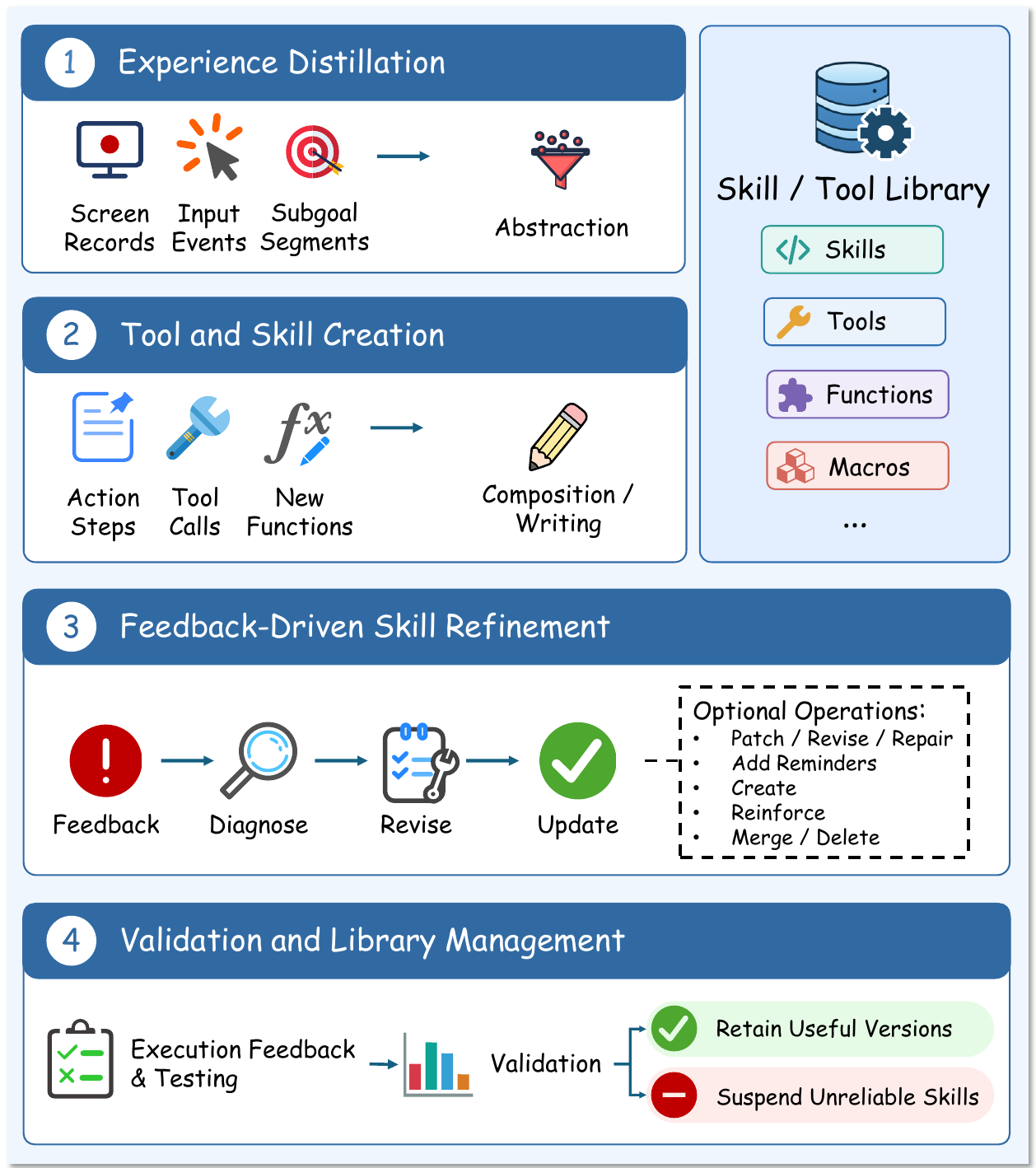}
  \vspace{-6mm}
  \SurveyFigureCaption{Skill and tool evolution across tasks}{Agents populate a shared library with skills distilled from task traces (1) and with new skills and tools built by composing action steps and tool calls or writing functions (2). Execution feedback helps agents diagnose shortcomings and refine skills (3). Validation uses execution feedback and tests to identify useful versions to retain and unreliable skills to suspend (4).}
  \vspace{-2mm}
  \label{fig:skill-tool-evolution}
\end{figure}

Agents can distill reusable skills from demonstrations and prior interactions or develop new skills and tools when existing resources fall short. As Fig.~\ref{fig:skill-tool-evolution} illustrates, storing these resources in a library enables reuse across tasks. Feedback from subsequent executions can then guide revisions, while validation helps determine which versions to retain.

\noindent$\bullet$\ \textbf{Experience Distillation.} A task trace can contain a procedure useful beyond the instance that produced it. Distilling that procedure provides guidance the agent can retrieve later or incorporate into policy training. Work2Skill~\cite{huang2026work2skill} derives parameterized GUI guidance by aligning human screen recordings, input events, and accessibility evidence, with static checks for structural consistency and provenance during induction. ArtisanCAD~\cite{xu2026artisancad} distills expert CAD recordings, macros, and notes into parameterized skills with operation dependencies and verification rules for later execution through CAD tools with multi-view visual feedback. MineEvolve~\cite{xie2026mineevolve} converts successful subgoal segments identified from visual observations into procedures with explicit triggers and checks, while repeated failures inform contextual remedies. SkillLens~\cite{liu2026skilllens} represents GUI procedures and interface views as visual skill cards and distills teacher behavior conditioned on these cards into a student policy that acts without retrieval. Skill-Guided Continuation Distillation~\cite{fan2026skill} extracts continuation guidance from GUI trajectories and uses skill-guided recoveries from policy-visited states to train the action model.

\noindent$\bullet$\ \textbf{Tool and Skill Creation.} Recurring action sequences or gaps in the available tools can motivate new executable procedures. Saving those procedures in a library makes them available to later attempts. Transductive Visual Programming~\cite{ICLR2026Transductive} abstracts recurring solved spatial-reasoning programs into parameterized library functions. META~\cite{CVPR2026META} derives reusable macro-tools from recurring long-video tool trajectories. During training, MACRO~\cite{fan2026evolving} registers frequent composite tool sequences from medical-imaging rollouts whose final answers match ground truth, making them available to subsequent rollouts. EchoPath~\cite{zhao2026echopath} converts validated GUI trajectories into callable memories with parameters, application-state preconditions, visual targets, and validation provenance. ReSight~\cite{che2026resight} distills exploration of robot-observation memories into executable builders and query tools that can be mounted for later tasks. VideoResearcher~\cite{ye2026videoresearcher} implements perception tools in response to long-video reasoning failures and retains tested designs for later evidence acquisition. Agent S4~\cite{agashe2026evolving} builds a persistent repository of executable computer-use interfaces and usage knowledge around a fixed task model and harness.

\noindent$\bullet$\ \textbf{Feedback-Driven Skill Refinement.} Reusing a skill can reveal situations its current procedure handles poorly, giving the agent feedback on what to revise and retain for later tasks. VisualSkill~\cite{jiang2026visualskill} analyzes failed training trajectories to identify underrepresented UI states, explores the corresponding interfaces, and patches persistent topic guides with the resulting visual evidence. ComfyClaw~\cite{li2026comfyclaw} clusters image-generation traces to propose revisions to its skill library. In EmbodiedBench~\cite{ICML2025EmbodiedBench} tasks with visual observations, EmbodiSkill~\cite{ju2026embodiskill} distinguishes procedural defects from execution lapses, revising the retained skill body for the former while appending reminders for the latter. EvoSkill-GUI~\cite{chen2026reflect} revises multi-file GUI skills using immediate execution feedback and separate failed-trajectory diagnosis. Persistent Skills~\cite{hu2026interaction} uses desktop trajectories and evaluator feedback to update a versioned library between iterations, with each iteration executing against a frozen snapshot. Designer-RSI~\cite{du2026designer} uses recurring user requests and graded executions to acquire or repair design procedures under fixed proposal and matched replay procedures, retaining updated skills for later design tasks.

\noindent$\bullet$\ \textbf{Validation and Library Management.} The library's contents can change as tests and execution feedback reveal when a skill or tool works reliably. These results can guide which versions to adopt or suspend. Transductive Visual Programming~\cite{ICLR2026Transductive} rewrites existing programs to use a candidate function, runs them, and checks the resulting answers before accepting the function. ComfyClaw~\cite{li2026comfyclaw} compares revised and existing libraries on synthesized held-out prompts and keeps revisions only when they show no measured performance drop. Reflect After Failure~\cite{li2026reflect} checks the reliability and evidence validity of procedures alongside task outcomes and efficiency before suspending unreliable skills. Agent S4~\cite{agashe2026evolving} separates construction trajectories from validation tasks that the interface-building agent cannot see, then uses validation performance to select the repository to deploy. VisualEvolve~\cite{wang2026visualevolve} screens revised visual-operation libraries for performance drops relative to saved versions on a fixed held-out set.

\subsection{Orchestration-Level Evolution}
\label{sec:l2-workflows}

When execution traces reveal gaps in the guidance or visual evidence supplied to the model, the agent can revise its prompts and context-construction procedures. Repeated failures at workflow handoffs can likewise motivate changes to routing, verification, and recovery rules. The agent retains these revisions in the harness for later tasks or trials.

\noindent$\bullet$\ \textbf{Prompt and Instruction Revision.} When task outcomes reveal shortcomings in what the model receives, the harness can revise the guidance it supplies or the procedures that assemble its evidence. Retaining the revised context rules changes what the model sees on later attempts. EvoGen-Harness~\cite{luo2026evogen} revises visual requirements and constraint priorities for frozen image generators, comparing guidance versions under matched stochastic generation conditions and checking for unwanted changes on held-out cases. GeoForge~\cite{xiao2026geoforge} turns Earth-observation task traces into reusable corrections and procedural constraints, excluding unsupported workflows and answer shortcuts tied to individual samples. In visual embodied tasks, SHAPER~\cite{wang2026self} uses rollout critiques to revise procedural guidance and then context-construction code, retaining versions selected by validation for held-out execution. EnactOpt~\cite{feng2026enactopt} uses action feedback to jointly revise execution guidance and how evidence is assembled for the planner. In video reasoning, revising the programs that select and process observations can change what visual evidence reaches the model. VideoHarness-RSI~\cite{xu2026videoharness} uses a fixed proposer and a development-accuracy criterion to search for programs that retrieve and pack evidence or invoke auxiliary calls around a frozen VLM, retaining accepted programs for later video question answering and further refinement. Video-RSI~\cite{luo2026video} uses a fixed Editor to revisit training videos and diagnose failures before revising observation processing and memory, then applies a fixed rule based on answer accuracy and visual cost to select revisions for later reuse and refinement.

\noindent$\bullet$\ \textbf{Workflow and Control Evolution.} A pattern in execution outcomes can reveal where the harness needs a different rule for taking the next step. Retaining such changes can affect individual decisions or reorganize a larger part of the workflow on later attempts. VEvo-Harness~\cite{wang2026vevo} updates operation scores from rendered outcomes and retains verified repair patterns for later rewriting, resampling, and tuning, while keeping its routing formula and acceptance thresholds fixed. AgentCanvas~\cite{zhou2026automating} revises typed execution graphs from simulator outcomes and logs, but held-out reversals and the exclusion of an evaluator-leaking candidate limit the strength of its generalization claims. InstAgent~\cite{luo2026instagent} retains revised behavior trees with visual checks and recovery branches at inter-robot handoffs following delivery failures. RegenHarness~\cite{wang2026regenharness} specifies a protocol that proposes changes to context rules, task templates, skill routing, and recovery policies from robot mission records, with fixed regression checks and release authorization governing adoption for later missions. LVMH~\cite{che2026lvmh} searches for executable memory policies that revise schemas, retrieval, and visual-evidence routing while retaining development records for later revisions. Continual Harness~\cite{karten2026continual} creates, revises, and removes sub-agents and skills from gameplay traces for later bootstrap runs. EvoGen-Harness~\cite{luo2026evogen} revises routing, verification, retry, and termination rules around a fixed deterministic base controller. SERA~\cite{du2026from} retains planning-context and control revisions selected from rollouts while using a fixed VLA executor. Evaluation can help determine which revisions to retain for later tasks. Zetta~\cite{ding2026zetta} keeps its action-selection model and orchestrator fixed while revising critics, recovery playbooks, and tools, and validates the resulting changes through diagnostic replay, fresh closed-loop execution, and multi-seed consolidation. GUI-HARVEST~\cite{yang2026gui} uses repeated failures grounded in action traces and before-and-after screenshots to generate runtime-code revisions, checks scores and predicted behavior before adoption, and transfers the frozen revised harness to a second desktop benchmark. Scale and Selection~\cite{wei2026scale} adopts a revised harness only when it outperforms the current harness on the same training rollout cases. KPHarness~\cite{chen2026kpharness} selects probes using a fixed rule inspired by upper confidence bounds and proposes gameplay-harness revisions from accumulated probe utilities and visit counts, with paired validation determining which edits to retain. HarnessMM-V~\cite{jiang2026harnessmmv} uses task success and token costs to revise how a visual harness authors, retains, and dispatches executable decisions, with a fixed procedure for proposing and selecting rules for reuse.

\subsection{Dynamic Data and Task Evolution}
\label{sec:l2-experience}

A visual agent's learning from its own attempts depends on the situations it encounters and the supervision available in each. The resulting trajectories and outcomes can guide the collection, synthesis, or refinement of training examples. As the agent's capabilities develop, its performance can guide the construction or prioritization of practice tasks. These tasks, in turn, shape the experience available in later learning rounds.

\noindent$\bullet$\ \textbf{Data Acquisition and Refinement.} What the agent encounters during execution can guide which examples it collects or refines for later training. CorrectAD~\cite{AAAI2026CorrectAD} analyzes driving-planner failures, formulates targeted data requirements, and synthesizes videos aligned with requested 3D layouts for subsequent planner training. HATS~\cite{CVPR2026HATS} collects semantically ambiguous GUI trajectories and iteratively repairs alignment between instructions and execution, with refinement feedback guiding later exploration. GenEvolve~\cite{NeurIPS2026GenEvolve} stores differences between higher- and lower-reward image-generation trajectories as visual experience, supplying a privileged teacher view whose self-distillation targets are absorbed by the deployed student. In robotics, PLD~\cite{ICLR2026Selfimprovingb} uses residual-RL specialists to collect recovery demonstrations from states reached by the base VLA, then distills these data into the generalist. SILVR~\cite{ICLR2026Selfimprovinga} accumulates successful rollouts for visual-planner fine-tuning, using human success judgments to filter examples in its main real-robot experiments. Scene2Demo~\cite{refresh_scene2demo2026} critiques multi-view simulated rollouts and repairs action flows before collecting offline demonstrations that train behavior-cloning policies. RECALL~\cite{karli2026recall} requests expert recovery demonstrations at states where the VLA is uncertain and uses them for fine-tuning, improving acquisition efficiency over passive demonstrations in a simulator-reset protocol on LIBERO-10~\cite{NeurIPS2023LIBERO}.

\noindent$\bullet$\ \textbf{Task and Curriculum Adaptation.} As an agent's capabilities grow, current task performance can identify areas for further practice and guide later task construction or prioritization. SEAgent~\cite{added_sun2026seagent} updates a software guidebook from step-level trajectory judgments and descriptions of GUI state changes, then applies the same curriculum generator and prompt template across phases to turn the guidebook and new feedback into tasks for subsequent actor training. ACuRL~\cite{NeurIPS2026Autonomous} synthesizes tasks suited to the current action-selection model from environment exploration and feedback from earlier training iterations. EvoCUA~\cite{NeurIPS2026EvoCUA} co-generates resource-grounded instructions and executable validators, turning verified rollouts and diagnosed failures into supervision for later controller training. SEES~\cite{li2026sees} diagnoses bottleneck skills in long-horizon VLA execution and constructs targeted simulated reinforcement-learning tasks around encountered states to update shared adapters. FIND~\cite{fang2026find} uses rolling success histories to prioritize feasible, lower-success tasks from a predefined library under a fixed sampling formula, training a residual controller around a frozen VLA. SimWorld Studio~\cite{kang2026simworld} raises navigation-task difficulty at a preset success-rate threshold, improving performance over fixed-difficulty training.

\section{L3 Toward Recursive Self-Improving Visual Agents}
\label{sec:level3}

A \emph{visual agent} can identify where it needs to improve by gathering new evidence and relating it to experience retained from earlier attempts. After the agent acts on those findings, evaluating the outcome can expose weaknesses in how the update was produced and guide revisions to the mechanism before it is used to produce another update. Table~\ref{tab:recursive-update-roles} shows how visual agents reuse these revisions.

\begin{table*}[!t]
  \centering
  \begingroup
  \SurveyTableStyle
  \colorlet{SurveyTableSection}{SurveyTableSection!70!white}
  \SurveyTableCaption{Update Targets and Mechanisms Supporting RSI in Visual Agents.}{tab:recursive-update-roles}
  \vspace{-2mm}
  \setlength{\tabcolsep}{.5pt}
  \newcolumntype{Q}{>{\centering\arraybackslash}p{24pt}}
  \newcommand{\RecursiveWork}[2]{#1 \cite{#2} & \TableWorkLinks{#2}}
  \newcommand{\RecursivePanel}[1]{%
    \SurveyTablePanel{6}{#1}}
  \newcommand{\RecursiveTableContents}{%
    \SurveyTableTopRule
    \SurveyTableRowColor{SurveyTableHeader}
    \textbf{Work} & \textbf{Links} & \textbf{Evolving Component} &
    \textbf{Update Method} & \textbf{Driving Signal} &
    \textbf{Recursive Role} \\
    \SurveyTableHeaderRule
    \RecursivePanel{(a) Evolving Learning Signals and Experience}
    \SurveyTableRowColor{white}
    \RecursiveWork{GUI0}{ACL2026GUI0}
    & Curriculum Agent
    & RL (GRPO)
    & Rollout Success / Task Length
    & Generate Actor-Training Tasks \\
    \SurveyTableRowColor{SurveyTableStripe}
    \RecursiveWork{VLAW}{added_guo2026vlaw}
    & Video World Model
    & Diffusion Fine-Tuning
    & Real-Robot Rollouts
    & Generate VLA-Training Data \\
    \SurveyTableRowColor{white}
    \RecursiveWork{Grounded-Dreamer}{jiang2026grounded}
    & World Model / Inverse Dynamics
    & Post-Training
    & Simulated Robot Trajectories
    & Supply Replay-Verified Data \\
    \SurveyTableRowColor{SurveyTableStripe}
    \RecursiveWork{PhysEvo}{tian2026physevo}
    & Trajectory-Review Tool
    & Tool Creation
    & Robot Rollout Records
    & Guide Later Skill Revisions \\
    \RecursivePanel{(b) Evolving Memory and Context Management}
    \SurveyTableRowColor{white}
    \RecursiveWork{EmbodiedRSI}{song2026embodiedrsi}
    & Memory Manager
    & RL (PPO)
    & Success Gain / Entropy Drop / Cost
    & Guide Experiments / Skill Updates \\
    \SurveyTableRowColor{SurveyTableStripe}
    \RecursiveWork{PRACTICE}{bai2026practice}
    & Skill Maintenance Policy
    & SFT / On-Policy Distillation
    & Executor Traces / Teacher Targets
    & Generate Skill-Library Edits \\
    \SurveyTableRowColor{white}
    \RecursiveWork{H\textsuperscript{2}-EMV}{baermann2026forget}
    & Memory Relevance Rules
    & Rule Revision
    & Feedback on Forgotten Details
    & Guide Later Summaries / Retention \\
    \SurveyTableRowColor{SurveyTableStripe}
    \RecursiveWork{VideoEvolve}{xu2026videoevolve}
    & Memory / Retrieval Policies
    & Alternating RL
    & Video QA Outcomes
    & Train Retriever on Rebuilt Memory \\
    \SurveyTableRowColor{white}
    \RecursiveWork{TaskMem}{zou2026taskmem}
    & Memory Writer Adapter
    & DPO
    & Recent Task Questions
    & Write Memory for Later Questions \\
    \RecursivePanel{(c) Evolving Update Mechanisms}
    \SurveyTableRowColor{white}
    \RecursiveWork{VideoEvoHarness}{xu2026videoevoharness}
    & Questioner
    & Rule Revision
    & Quality Checks / Solver Traces
    & Supply Harness-Update Tasks \\
    \SurveyTableRowColor{SurveyTableStripe}
    \RecursiveWork{CoDA}{yu2026coda}
    & Generation / Training Programs
    & Program Compilation
    & Rollout Diagnoses / Policy Progress
    & Generate Data / Guide Training \\
    \SurveyTableRowColor{white}
    \RecursiveWork{AdaRSI}{wang2026adarsi}
    & Shared Development Experience
    & Round-End Summarization
    & Development / Validation Records
    & Guide Diagnosis / Data Collection \\
    \SurveyTableRowColor{SurveyTableStripe}
    \RecursiveWork{HarnessPAI}{wang2026harnesspai}
    & Repair Skill Library
    & Skill Distillation
    & Rollout Failures / Successful Fixes
    & Guide Later Program Revisions \\
    \SurveyTableRowColor{white}
    \RecursiveWork{RecurIR}{gao2026recurir}
    & Task Generator / Revision Module
    & Rule Tests / Revision Learning
    & Degradation Tests / Plan Outcomes
    & Generate Tests / Guide Revisions \\
    \RecursivePanel{(d) Evolving Evaluation and Validation}
    \SurveyTableRowColor{white}
    \RecursiveWork{Co-EPG}{AAAI2026CoEPG}
    & Grounding Model
    & SFT on Verified Data
    & Verified Plans / Action Labels
    & Supply Planner Rewards / Data \\
    \SurveyTableRowColor{SurveyTableStripe}
    \RecursiveWork{Agent0-VL}{ICML2026Agent0VL}
    & Shared Solver--Verifier
    & SFT / RL (GRPO)
    & SFT Traces / Composite Rewards
    & Supply Step-Level Rewards \\
    \SurveyTableRowColor{white}
    \RecursiveWork{VeriCUA}{wang2026vericua}
    & Judge Policy
    & SFT / RL (REINFORCE++)
    & Outcome Labels / Probe Use
    & Reward Actor with Frozen Verifier \\
    \SurveyTableRowColor{SurveyTableStripe}
    \RecursiveWork{RewardHarness}{zhang2026rewardharness}
    & Evaluation Skills / Tools
    & Validation-Gated Editing
    & Human Preference-Ranking Errors
    & Reward Image-Editor RL \\
    \SurveyTableRowColor{white}
    \RecursiveWork{SkillRubric}{jiang2026skillrubric}
    & Evaluation Rubrics
    & Diagnosis / Offline Validation
    & On-Policy Traces / Fixed Verdicts
    & Assign Turn-Level RL Credit \\
    \SurveyTableBottomRule
  }
  \newsavebox{\RecursiveTableBox}
  \sbox{\RecursiveTableBox}{\begin{tabular}{lQllll}\RecursiveTableContents\end{tabular}}
  \typeout{L3 Table: \the\wd\RecursiveTableBox; text: \the\linewidth}
  \ifdim\wd\RecursiveTableBox>\linewidth
    \PackageError{RecursiveUpdateTable}{Single-line table exceeds text width}%
    {Shorten a label without scaling the font.}
  \fi
  \begingroup
  \setlength{\tabcolsep}{\dimexpr\tabcolsep+(\linewidth-\wd\RecursiveTableBox)/12\relax}
  \begin{tabular}{lQllll}\RecursiveTableContents\end{tabular}
  \endgroup
  \SurveyTableDescription{%
    \TableLinksLegend\enspace
    \textit{Recursive Role} indicates how the revised component contributes to a later update.
    RL = reinforcement learning, SFT = supervised fine-tuning,
    GRPO = group relative policy optimization, PPO = proximal policy optimization,
    DPO = direct preference optimization, QA = question answering, and VLA = vision-language-action.
  }
  \endgroup
  \vspace{-6mm}
\end{table*}

\subsection{Evolving Learning Signals and Experience}
\label{sec:l3-learning}

As an agent develops new capabilities, earlier attempts may tell it less about the weaknesses that remain. Revising how it selects practice tasks or generates imagined trajectories can bring more informative experience into training. How the agent interprets the outcomes also matters, since diagnosing a failure can reveal which capability needs further revision.

GUI0~\cite{ACL2026GUI0} uses rewards derived from rollout success rates to train a curriculum agent to favor tasks near the actor's capability boundary, then uses the revised agent to generate the next round of actor-training tasks. Marvo~\cite{yang2026self} updates a visual Challenger from tool-use benefits and reuses it to generate later training questions. SCALE~\cite{CVPR2026Learning} fine-tunes an action Selector on misunderstood screenshot transitions and uses it to gather experience for later Predictor training under a fixed Judger. PhysEvo~\cite{tian2026physevo} reuses a trajectory-review tool in a later round to trace a packing failure to contact with raised flaps, informing a skill revision that requires clearance before turning. RIWANav~\cite{xie2026riwanav} refines its visual world model on a curriculum drawn from the navigation policy's action proposals and reuses its predictions to guide the next policy update. WMPO~\cite{ICLR2026Wmpo} alternates policy rollouts with world-model refinement in its StackThree experiment, using the revised model to generate imagined trajectories for the next policy update. VLAW~\cite{added_guo2026vlaw} refines its video world model with real-robot policy rollouts and reuses it to generate trajectories for subsequent VLA training. WoVR~\cite{jiang2026wovr} refines its action-conditioned world model on rollouts from the updated policy, then uses the revised model to generate imagined trajectories for another round of policy training. Grounded-Dreamer~\cite{jiang2026grounded} adapts a video world model and an inverse-dynamics model on simulator trajectories, then uses the revised models to synthesize trajectories that are filtered by simulation replay before offline training of real-robot policies.

\subsection{Evolving Memory and Context Management}
\label{sec:l3-memory}

A new problem can make earlier experience relevant again. The evidence available to the agent depends on which observations it retains, how it organizes them, and what it retrieves for the current context. Revising these procedures changes which evidence the agent retains or supplies for later experiments, skill revisions, and training.

EmbodiedRSI~\cite{song2026embodiedrsi} learns a memory manager from physical-experiment outcomes and reuses it to guide later experiments and skill revisions. PRACTICE~\cite{bai2026practice} learns a procedural-knowledge maintenance policy from visual embodied trajectories and reuses it to revise the skill library for a frozen executor. H\textsuperscript{2}-EMV~\cite{baermann2026forget} revises natural-language relevance rules in response to feedback on forgotten details and uses them to summarize later robot observations and decide which records to retain. AutoMMemo~\cite{sui2026learning} searches over executable memory-writing and retrieval programs, rebuilds memory after each revision, and evaluates the resulting guidance on separate tasks. VideoEvolve~\cite{xu2026videoevolve} updates its memory construction policy using question-answering outcomes from the current retriever, then rebuilds video memories with the revised policy to train the next retriever. In simulated streaming video tasks, TaskMem~\cite{zou2026taskmem} ranks candidate memories by relevance to recent questions and tunes its writer's adapter with direct preference optimization, then uses the revised writer to build memories for later questions.

\subsection{Evolving Update Mechanisms}
\label{sec:l3-update-mechanisms}

An update mechanism can use diagnoses to propose update tasks, design revisions, or implement changes. Update outcomes can expose weaknesses and guide revisions to the mechanism. The agent can then reuse the revised mechanism to design or apply more targeted updates.

VideoEvoHarness~\cite{xu2026videoevoharness} revises Questioner rules from video-question checks and Solver traces and reuses the Questioner to supply later harness-update tasks. CoDA~\cite{yu2026coda} revises scene-generation and policy-training programs from visual rollout diagnoses and learning progress for later data and policy updates. The Trinity~\cite{tang2026trinity} consolidates Creator modules revised after failed world verification and reuses them to build training worlds for later Explorer harness updates. RSIgen~\cite{liu2026rsigen} reuses updated understanding and generation roles to construct later curricula and supervision. AdaRSI~\cite{wang2026adarsi} distills earlier development records into guidance for diagnosing persistent failures and targeting VLA adaptation in later rounds. HarnessPAI~\cite{wang2026harnesspai} distills rollout failures and successful fixes into repair skills that its coding agent retrieves to revise programs for later tasks. RecurIR~\cite{gao2026recurir} updates its task generator and plan-revision module using tests under targeted degradation changes, then uses the revised generator to supply counterexamples for later restoration policy updates.

\subsection{Evolving Evaluation and Validation}
\label{sec:l3-closed-loop}

Evaluation shapes improvement by determining how task outcomes inform training and which revisions the agent adopts. As new cases expose gaps in these judgments, execution results and external assessments can guide revisions to what the evaluator checks and how it interprets the evidence. Once the revised evaluator guides later updates, independent task outcomes help establish whether the resulting changes improve the agent's capabilities.

Co-EPG~\cite{AAAI2026CoEPG} reuses updated GUI-grounding checkpoints to supply rewards and verified data for later planner training. Agent0-VL~\cite{ICML2026Agent0VL} updates a shared Solver--Verifier whose critiques provide process rewards for later visual reasoning training. CineEvolve~\cite{zhang2026cineevolve} reuses updated Critics to reward later Director updates and trains Critics on annotated Director rollouts. MagicGUI-RMS~\cite{li2026magicgui} retrains its domain-specific reward model on disagreements with a general-purpose evaluator and reuses the revised model to verify actions for subsequent GUI actor training. JarvisEvo~\cite{CVPR2026Jarvisevo} calibrates an image-editing evaluator against human scores before reusing it for editor training. RewardHarness~\cite{zhang2026rewardharness} revises evaluation skills and tools from preference-ranking errors and reuses them for later image-editor training. VISA~\cite{zeng2026visa} expands image-conditioned verifier bindings through sandbox validation and reuses them in later data synthesis and reinforcement-learning rewards. SkillRubric~\cite{jiang2026skillrubric} co-evolves actor guidance and screenshot-grounded rubrics before using the revised rubrics to supervise later policy updates. VeriCUA~\cite{wang2026vericua} learns what to check and when to stop from executable GUI task outcomes and probe-use feedback, then tests the frozen verifier as a reward source for subsequent actor training in matched runs. LIT~\cite{jian2026learning} revises video-quality criteria and inspection procedures from production traces to guide later Director skill updates, while its meta engineer also revises its own task-selection and revision skills for use in later evolution epochs.

\section{Benchmarks and Evaluation of Visual-Agent Evolution}
\label{sec:benchmarks}

\providecommand{\RaggedRight}{\raggedright}
\definecolor{DatasetLevelIIIFill}{HTML}{FAF4DD}
\definecolor{DatasetLevelIIIText}{HTML}{7B6B3E}
\definecolor{DatasetLevelIVFill}{HTML}{F8E6E6}
\definecolor{DatasetLevelIVText}{HTML}{8C5656}
\definecolor{DatasetLevelVFill}{HTML}{E6EEF8}
\definecolor{DatasetLevelVText}{HTML}{426578}
\definecolor{DatasetLinkPaper}{HTML}{3F78A8}
\definecolor{DatasetLinkCode}{HTML}{328873}
\definecolor{DatasetLinkData}{HTML}{C18B48}
\definecolor{DatasetStripe}{HTML}{F8F8F8}
\def\DatasetBadge#1#2#3{%
  \begingroup\setlength{\fboxsep}{0pt}%
  \colorbox{#1}{\parbox[c][7.8pt][c]{22pt}{%
      \centering\color{#2}\rmfamily\bfseries #3}}%
  \endgroup}
\newcommand{\DatasetCoverage}[2]{%
  \DatasetBadge{DatasetLevel#1Fill}{DatasetLevel#1Text}{#1-#2}}
\newcommand{\DatasetCell}[1]{\mbox{#1}}
\newcommand{\DatasetLinkSlot}[3]{%
  \makebox[1.1em][c]{%
    \if U#1%
      \textnormal{-}%
    \else
      \href{#2}{\raisebox{0.03ex}{#3}}%
    \fi}}
\def\DatasetLinksByStatus#1#2#3\relax#4#5#6{%
  \mbox{\DatasetLinkSlot{#1}{#4}{\SurveyPaperIcon}\hspace{0.12em}%
    \DatasetLinkSlot{#2}{#5}{\SurveyCodeIcon}\hspace{0.12em}%
    \DatasetLinkSlot{#3}{#6}{\SurveyDataIcon}}}
\newcommand{\DatasetLinks}[4][VVV]{\DatasetLinksByStatus#1\relax{#2}{#3}{#4}}
\def\DatasetTableStyle{%
  \SurveyTableStyle
  \setlength{\tabcolsep}{.5pt}%
  \colorlet{SurveyTableStripe}{DatasetStripe}%
  \colorlet{SurveyLinkPaper}{DatasetLinkPaper}%
  \colorlet{SurveyLinkCode}{DatasetLinkCode}%
  \colorlet{SurveyLinkData}{DatasetLinkData}%
  \arrayrulecolor{SurveyTableRule}}
\def\DatasetName#1{\DatasetCell{#1}}
\def\DatasetTopRule{\SurveyTableTopRule}
\def\DatasetHeader{%
  \SurveyTableRowColor{SurveyTableHeader}%
  \DatasetCell{\textbf{Benchmark}} & \DatasetCell{\textbf{Links}} & \DatasetCell{\textbf{Setting}} &
  \DatasetCell{\textbf{Evaluation Target}} & \DatasetCell{\textbf{Focus}} & \DatasetCell{\textbf{Scale}} &
  \DatasetCell{\textbf{Unit}} & \DatasetCell{\textbf{Coverage}} \\
  \SurveyTableHeaderRule}
\newcommand{\DatasetLegend}[2]{%
  \SurveyTableDescription{%
    \textit{Links}\enspace \SurveyPaperIcon\,Paper / Docs\quad
    \SurveyCodeIcon\,Code\quad \SurveyDataIcon\,Data / Evaluation Assets.
    - = Unverified Public Availability.
    \textit{Survey~Coverage}\enspace #1\space #2%
  }}
\newcommand{\DatasetTableCaption}[2]{%
  \SurveyTableCaption{#1}{#2}}

\newcommand{\VisualAgentTaskBenchmarkTable}{%
  \begin{table*}[t]
    \centering
    \begingroup
    \DatasetTableStyle
    \DatasetTableCaption{Benchmarks and Taxonomy Coverage for Task-Level Evaluation.}{tab:visual-agent-task-benchmarks}
    \vspace{-2mm}
    \begin{tabular}{w{l}{105.5pt}w{c}{36pt}w{l}{65pt}w{l}{78pt}w{l}{92.7pt}w{l}{28.2pt}w{l}{68pt}w{c}{34.6pt}}
      \DatasetTopRule
      \DatasetHeader
      \SurveyTableRowColor{white}
      \DatasetName{ActiView~\cite{ACL2025ActiView}}                     & \DatasetLinks[VVV]{https://aclanthology.org/2025.acl-long.376/}{https://github.com/THUNLP-MT/ActiView}{https://huggingface.co/datasets/wangphoebe/ActiView}                                                                                                            & \DatasetCell{Image Interaction} & \DatasetCell{Task Capability}   & \DatasetCell{Shifting / Zooming}          & \DatasetCell{325}               & \DatasetCell{Distinct Questions}   & \DatasetCell{\mbox{\DatasetCoverage{III}{A}}} \\
      \SurveyTableRowColor{SurveyTableStripe}
      \DatasetName{AgentVQA~\cite{gupta2026agentvqa}}                   & \DatasetLinks[VVV]{https://arxiv.org/abs/2610.31494}{https://github.com/zlab-princeton/AgentVQA}{https://huggingface.co/datasets/AgentVQA/AgentVQA}                                                                                                                    & \DatasetCell{Offline}           & \DatasetCell{Task Capability}   & \DatasetCell{Agent-State Understanding}   & \DatasetCell{1,000}             & \DatasetCell{Examples}             & \DatasetCell{\mbox{\DatasetCoverage{III}{A}}} \\
      \SurveyTableRowColor{white}
      \DatasetName{Video-MME-v2~\cite{fu2026video}}                     & \DatasetLinks[VVV]{https://arxiv.org/abs/2604.05015}{https://github.com/MME-Benchmarks/Video-MME-v2}{https://huggingface.co/datasets/MME-Benchmarks/Video-MME-v2}                                                                                                      & \DatasetCell{Offline}           & \DatasetCell{Task Capability}   & \DatasetCell{Temporal Consistency}        & \DatasetCell{3,200}             & \DatasetCell{QA Pairs}             & \DatasetCell{\mbox{\DatasetCoverage{III}{A}}} \\
      \SurveyTableRowColor{SurveyTableStripe}
      \DatasetName{ENACT~\cite{ICLR2026ENACT}}                          & \DatasetLinks[VVV]{https://arxiv.org/abs/2511.20937}{https://github.com/mll-lab-nu/ENACT}{https://huggingface.co/datasets/MLL-Lab/ENACT}                                                                                                                               & \DatasetCell{Offline}           & \DatasetCell{Task Capability}   & \DatasetCell{State--Action Reasoning}     & \DatasetCell{8,972}             & \DatasetCell{QA Pairs}             & \DatasetCell{\mbox{\DatasetCoverage{III}{A}}} \\
      \SurveyTableRowColor{white}
      \DatasetName{MPR-GUI-Bench~\cite{ACL2026MPRGUI}}                  & \DatasetLinks[VUV]{https://arxiv.org/abs/2512.00756}{}{https://huggingface.co/datasets/chenruihan/MPR-GUI-Bench}                                                                                                                                                       & \DatasetCell{Offline}           & \DatasetCell{Task Capability}   & \DatasetCell{Multilingual GUI Reasoning}  & \DatasetCell{12,936}            & \DatasetCell{Samples}              & \DatasetCell{\mbox{\DatasetCoverage{III}{A}}} \\
      \SurveyTableRowColor{SurveyTableStripe}
      \DatasetName{ActiveArena-Bench~\cite{li2026activearena}}          & \DatasetLinks[VVP]{https://arxiv.org/abs/2609.24124v2}{https://github.com/leeibo/ActiveArena}{https://huggingface.co/datasets/leeibo/ActiveArena-Data}                                                                                                                 & \DatasetCell{Simulation}        & \DatasetCell{Task Capability}   & \DatasetCell{Active Visual Evidence}      & \DatasetCell{35}                & \DatasetCell{Tasks}                & \DatasetCell{\mbox{\DatasetCoverage{III}{A}}} \\
      \SurveyTableRowColor{white}
      \DatasetName{OpenEQA~\cite{CVPR2024OpenEQA}}                      & \DatasetLinks[VPV]{https://openaccess.thecvf.com/content/CVPR2024/papers/Majumdar_OpenEQA_Embodied_Question_Answering_in_the_Era_of_Foundation_Models_CVPR_2024_paper.pdf}{https://github.com/facebookresearch/open-eqa}{https://github.com/facebookresearch/open-eqa} & \DatasetCell{Simulation}        & \DatasetCell{Task Capability}   & \DatasetCell{Exploration for Active QA}   & \DatasetCell{557}               & \DatasetCell{HM3D Questions}       & \DatasetCell{\mbox{\DatasetCoverage{III}{A}}} \\
      \SurveyTableRowColor{SurveyTableStripe}
      \DatasetName{OSWorld 2.0~\cite{yuan2026osworld2}}                 & \DatasetLinks[VVV]{https://arxiv.org/abs/2606.29537}{https://github.com/xlang-ai/OSWorld-V2/tree/v2026.06.24}{https://huggingface.co/datasets/xlangai/osworld_v2_tasks}                                                                                                & \DatasetCell{Interactive GUI}   & \DatasetCell{Task Capability}   & \DatasetCell{Long-Horizon Execution}      & \DatasetCell{108}               & \DatasetCell{Workflows}            & \DatasetCell{\mbox{\DatasetCoverage{III}{B}}} \\
      \SurveyTableRowColor{white}
      \DatasetName{WebArena-Pro~\cite{NeurIPS2026WebArenaPro}}          & \DatasetLinks[VVV]{https://arxiv.org/abs/2611.17955}{https://github.com/McGill-NLP/webarena-pro}{https://huggingface.co/WebArena-Pro}                                                                                                                                  & \DatasetCell{Interactive GUI}   & \DatasetCell{Task Capability}   & \DatasetCell{Multimodal Web Tasks}        & \DatasetCell{300}               & \DatasetCell{Test Tasks}           & \DatasetCell{\mbox{\DatasetCoverage{III}{B}}} \\
      \SurveyTableRowColor{SurveyTableStripe}
      \DatasetName{Ego2Web~\cite{CVPR2026Ego2Web}}                      & \DatasetLinks[VVV]{https://arxiv.org/abs/2603.22529}{https://github.com/Yui010206/Ego2Web}{https://huggingface.co/datasets/Shoubin/Ego2Web}                                                                                                                            & \DatasetCell{Interactive GUI}   & \DatasetCell{Task Capability}   & \DatasetCell{Egocentric Web Tasks}        & \DatasetCell{500}               & \DatasetCell{Video-Task Pairs}     & \DatasetCell{\mbox{\DatasetCoverage{III}{B}}} \\
      \SurveyTableRowColor{white}
      \DatasetName{NaturalGAIA~\cite{ACL2026NaturalGAIA}}               & \DatasetLinks[VVU]{https://aclanthology.org/2026.acl-long.2207/}{https://github.com/KeLes-Coding/NatureGAIA}{}                                                                                                                                                         & \DatasetCell{Interactive GUI}   & \DatasetCell{Task Capability}   & \DatasetCell{User-Intent Satisfaction}    & \DatasetCell{276}               & \DatasetCell{Tasks}                & \DatasetCell{\mbox{\DatasetCoverage{III}{B}}} \\
      \SurveyTableRowColor{SurveyTableStripe}
      \DatasetName{ProSoftArena~\cite{CVPR2026ProSoftArena}}            & \DatasetLinks[VVV]{https://arxiv.org/abs/2601.02399}{https://github.com/jiaxin-ai/ProSoftArena}{https://github.com/jiaxin-ai/ProSoftArena}                                                                                                                             & \DatasetCell{Interactive GUI}   & \DatasetCell{Task Capability}   & \DatasetCell{Professional Software}       & \DatasetCell{436}               & \DatasetCell{Tasks}                & \DatasetCell{\mbox{\DatasetCoverage{III}{B}}} \\
      \SurveyTableRowColor{white}
      \DatasetName{SuperAPP Benchmark~\cite{ACL2026GUI0}}               & \DatasetLinks[VUU]{https://aclanthology.org/2026.acl-long.2052/}{}{}                                                                                                                                                                                                   & \DatasetCell{Interactive GUI}   & \DatasetCell{Task Capability}   & \DatasetCell{Super-App Task Execution}    & \DatasetCell{230}               & \DatasetCell{Tasks}                & \DatasetCell{\mbox{\DatasetCoverage{III}{B}}} \\
      \SurveyTableRowColor{SurveyTableStripe}
      \DatasetName{VisualWebArena~\cite{ACL2024VisualWebArena}}         & \DatasetLinks[VVV]{https://aclanthology.org/2024.acl-long.50/}{https://github.com/web-arena-x/visualwebarena}{https://github.com/web-arena-x/visualwebarena/tree/main/config_files}                                                                                    & \DatasetCell{Interactive GUI}   & \DatasetCell{Task Capability}   & \DatasetCell{Visually Grounded Web Tasks} & \DatasetCell{910}               & \DatasetCell{Tasks}                & \DatasetCell{\mbox{\DatasetCoverage{III}{B}}} \\
      \SurveyTableRowColor{white}
      \DatasetName{MemU-Bench~\cite{liu2026memubench}}                  & \DatasetLinks[VVV]{https://arxiv.org/abs/2611.00191}{https://github.com/ZJU-REAL/MemU-Bench}{https://huggingface.co/datasets/vagrt/MemU-Bench-data}                                                                                                                    & \DatasetCell{Mobile GUI}        & \DatasetCell{Task Capability}   & \DatasetCell{Execution from User History} & \DatasetCell{118}               & \DatasetCell{GUI Tasks}            & \DatasetCell{\mbox{\DatasetCoverage{III}{B}}} \\
      \SurveyTableRowColor{SurveyTableStripe}
      \DatasetName{AsgardBench~\cite{tupini2026asgardbench}}            & \DatasetLinks[VVV]{https://arxiv.org/abs/2603.15888v2}{https://github.com/microsoft/AsgardBench}{https://github.com/microsoft/AsgardBench/tree/main/Generated}                                                                                                         & \DatasetCell{Simulation}        & \DatasetCell{Task Capability}   & \DatasetCell{Visual Plan Adaptation}      & \DatasetCell{108}               & \DatasetCell{Task Instances}       & \DatasetCell{\mbox{\DatasetCoverage{III}{B}}} \\
      \SurveyTableRowColor{white}
      \DatasetName{DeepPHY~\cite{AAAI2026DeepPHY}}                      & \DatasetLinks[VVU]{https://arxiv.org/abs/2508.05405}{https://github.com/XinrunXu/DeepPHY}{}                                                                                                                                                                            & \DatasetCell{Simulation}        & \DatasetCell{Task Capability}   & \DatasetCell{Physical Planning / Action}  & \DatasetCell{6}                 & \DatasetCell{Envs.}                & \DatasetCell{\mbox{\DatasetCoverage{III}{B}}} \\
      \SurveyTableRowColor{SurveyTableStripe}
      \DatasetName{PokeAgent (Speedrunning)~\cite{karten2026pokeagent}} & \DatasetLinks[VVU]{https://arxiv.org/abs/2603.15563v2}{https://github.com/sethkarten/continual-harness}{}                                                                                                                                                              & \DatasetCell{Simulation}        & \DatasetCell{Task Capability}   & \DatasetCell{Long-Horizon RPG Planning}   & \DatasetCell{15}                & \DatasetCell{First-Gym Milestones} & \DatasetCell{\mbox{\DatasetCoverage{III}{B}}} \\
      \SurveyTableRowColor{white}
      \DatasetName{ALFRED~\cite{CVPR2020ALFRED}}                        & \DatasetLinks[VVV]{https://arxiv.org/abs/1912.01734}{https://github.com/askforalfred/alfred}{https://github.com/askforalfred/alfred/tree/master/data}                                                                                                                  & \DatasetCell{Simulation}        & \DatasetCell{Task Capability}   & \DatasetCell{Household Goal Completion}   & \DatasetCell{8,055}             & \DatasetCell{Expert Demos}         & \DatasetCell{\mbox{\DatasetCoverage{III}{B}}} \\
      \SurveyTableRowColor{SurveyTableStripe}
      \DatasetName{OSWorld-MCP~\cite{ICLR2026OSWorldMCP}}               & \DatasetLinks[VVV]{https://arxiv.org/abs/2510.24563}{https://github.com/X-PLUG/OSWorld-MCP}{https://github.com/X-PLUG/OSWorld-MCP/tree/main/evaluation_examples}                                                                                                       & \DatasetCell{Interactive GUI}   & \DatasetCell{Task Capability}   & \DatasetCell{Tool Invocation / Outcomes}  & \DatasetCell{361}               & \DatasetCell{Evaluated Tasks}      & \DatasetCell{\mbox{\DatasetCoverage{III}{C}}} \\
      \SurveyTableRowColor{white}
      \DatasetName{AgentVista~\cite{NeurIPS2026AgentVista}}             & \DatasetLinks[VVV]{https://arxiv.org/abs/2602.23166}{https://github.com/hkust-nlp/AgentVista}{https://huggingface.co/datasets/Warrieryes/AgentVista}                                                                                                                   & \DatasetCell{Tool Environment}  & \DatasetCell{Task Capability}   & \DatasetCell{Long Tool Workflows}         & \DatasetCell{209}               & \DatasetCell{Tasks}                & \DatasetCell{\mbox{\DatasetCoverage{III}{C}}} \\
      \SurveyTableRowColor{SurveyTableStripe}
      \DatasetName{EgoBench~\cite{liu2026egobench}}                     & \DatasetLinks[VVV]{https://arxiv.org/abs/2605.27820v2}{https://github.com/Wind-Flipped/EgoBench}{https://huggingface.co/datasets/emodiary/EgoBench}                                                                                                                    & \DatasetCell{Tool Environment}  & \DatasetCell{Task Capability}   & \DatasetCell{Video / Tool / User Goals}   & \DatasetCell{1,590}             & \DatasetCell{Tasks}                & \DatasetCell{\mbox{\DatasetCoverage{III}{C}}} \\
      \SurveyTableRowColor{white}
      \DatasetName{3DCodeBench~\cite{NeurIPS20263DCodeBench}}           & \DatasetLinks[VVV]{https://arxiv.org/abs/2606.01057}{https://github.com/gaoypeng/3dcodebench}{https://huggingface.co/datasets/YipengGao/3DCodeBench}                                                                                                                   & \DatasetCell{3D Modeling}       & \DatasetCell{Task Capability}   & \DatasetCell{3D Code Refinement}          & \DatasetCell{212}               & \DatasetCell{Object Categories}    & \DatasetCell{\mbox{\DatasetCoverage{III}{D}}} \\
      \SurveyTableRowColor{SurveyTableStripe}
      \DatasetName{3DHarnessBench~\cite{liu20263dharnessbench}}         & \DatasetLinks[VVV]{https://arxiv.org/abs/2609.06535}{https://github.com/llada60/3DHarnessBench}{https://huggingface.co/datasets/lingada/3DHarnessBench}                                                                                                                & \DatasetCell{3D Modeling}       & \DatasetCell{Task Capability}   & \DatasetCell{View / Geometry Access}      & \DatasetCell{100}               & \DatasetCell{Objects}              & \DatasetCell{\mbox{\DatasetCoverage{III}{D}}} \\
      \SurveyTableRowColor{white}
      \DatasetName{PPT-Eval~\cite{ICML2026PPTEval}}                     & \DatasetLinks[VVV]{https://arxiv.org/abs/2606.31154}{https://github.com/microsoft/ppteval}{https://github.com/microsoft/ppteval/tree/main/task_registry}                                                                                                               & \DatasetCell{Interactive GUI}   & \DatasetCell{Task Capability}   & \DatasetCell{Slide Creation / Editing}    & \DatasetCell{120}               & \DatasetCell{Tasks}                & \DatasetCell{\mbox{\DatasetCoverage{III}{D}}} \\
      \SurveyTableRowColor{SurveyTableStripe}
      \DatasetName{WeGenEditBench~\cite{he2026evaluation}}              & \DatasetLinks[VUU]{https://arxiv.org/abs/2610.41210}{}{}                                                                                                                                                                                                               & \DatasetCell{Offline}           & \DatasetCell{Task Capability}   & \DatasetCell{Editing Constraint Checks}   & \DatasetCell{2,740}             & \DatasetCell{Eval. Instances}      & \DatasetCell{\mbox{\DatasetCoverage{III}{D}}} \\
      \SurveyTableRowColor{white}
      \DatasetName{RefCraft~\cite{wang2026refcraft}}                    & \DatasetLinks[VVV]{https://arxiv.org/abs/2610.26826v2}{https://github.com/wangyuchi369/RefCraft}{https://huggingface.co/datasets/YuchiWang/RefCraft}                                                                                                                   & \DatasetCell{Tool Environment}  & \DatasetCell{Task Capability}   & \DatasetCell{Reference-Guided Webpages}   & \DatasetCell{300}               & \DatasetCell{Tasks}                & \DatasetCell{\mbox{\DatasetCoverage{III}{D}}} \\
      \SurveyTableRowColor{SurveyTableStripe}
      \DatasetName{GenEvolve-Bench~\cite{NeurIPS2026GenEvolve}}         & \DatasetLinks[VVV]{https://arxiv.org/abs/2605.21605}{https://github.com/MeiGen-AI/GenEvolve}{https://huggingface.co/datasets/MeiGen-AI/GenEvolve-Data-Bench}                                                                                                           & \DatasetCell{Tool Environment}  & \DatasetCell{Task Capability}   & \DatasetCell{Tool-Assisted Generation}    & \DatasetCell{594}               & \DatasetCell{Held-Out Prompts}     & \DatasetCell{\mbox{\DatasetCoverage{III}{D}}} \\
      \SurveyTableRowColor{white}
      \DatasetName{ComfyBench~\cite{CVPR2025ComfyBench}}                & \DatasetLinks[VVV]{https://arxiv.org/abs/2409.01392}{https://github.com/xxyQwQ/ComfyBench}{https://github.com/xxyQwQ/ComfyBench/tree/main/dataset}                                                                                                                     & \DatasetCell{Tool Environment}  & \DatasetCell{Task Capability}   & \DatasetCell{Visual Workflow Outputs}     & \DatasetCell{200}               & \DatasetCell{Tasks}                & \DatasetCell{\mbox{\DatasetCoverage{III}{D}}} \\
      \SurveyTableRowColor{SurveyTableStripe}
      \DatasetName{VAB-CSS~\cite{ICLR2025VisualAgentBench}}             & \DatasetLinks[VVU]{https://proceedings.iclr.cc/paper_files/paper/2025/hash/eea71dc576381b88f2a0ca4dedc2140d-Abstract-Conference.html}{https://github.com/THUDM/VisualAgentBench}{}                                                                                     & \DatasetCell{Tool Environment}  & \DatasetCell{Task Capability}   & \DatasetCell{Target Screenshot Match}     & \DatasetCell{165}               & \DatasetCell{Test Cases}           & \DatasetCell{\mbox{\DatasetCoverage{III}{D}}} \\
      \SurveyTableRowColor{white}
      \DatasetName{VPI-Bench~\cite{ICLR2026VPIBench}}                   & \DatasetLinks[VVV]{https://arxiv.org/abs/2506.02456}{https://github.com/cua-framework/agents}{https://huggingface.co/datasets/VPI-Bench/vpi-bench}                                                                                                                     & \DatasetCell{Interactive GUI}   & \DatasetCell{Task Capability}   & \DatasetCell{Visual Prompt Injection}     & \DatasetCell{306}               & \DatasetCell{Test Cases}           & \DatasetCell{\mbox{\DatasetCoverage{III}{E}}} \\
      \SurveyTableRowColor{SurveyTableStripe}
      \DatasetName{CUA-Recovery~\cite{hu2026cuarecovery}}               & \DatasetLinks[VVU]{https://arxiv.org/abs/2610.27645}{https://github.com/xu-hu-2002/CUA-Recovery}{}                                                                                                                                                                     & \DatasetCell{Interactive GUI}   & \DatasetCell{Task Capability}   & \DatasetCell{Inherited-Error Recovery}    & \DatasetCell{2,250}             & \DatasetCell{Error States}         & \DatasetCell{\mbox{\DatasetCoverage{III}{E}}} \\
      \SurveyTableRowColor{white}
      \DatasetName{ProBench~\cite{AAAI2026Probench}}                    & \DatasetLinks[VUU]{https://arxiv.org/abs/2511.09157}{}{}                                                                                                                                                                                                               & \DatasetCell{Mobile GUI}        & \DatasetCell{Task Capability}   & \DatasetCell{Process-Sensitive Outcomes}  & \DatasetCell{\ensuremath{>200}} & \DatasetCell{Tasks}                & \DatasetCell{\mbox{\DatasetCoverage{III}{E}}} \\
      \SurveyTableRowColor{SurveyTableStripe}
      \DatasetName{FailBench~\cite{navasardyan2026failbench}}           & \DatasetLinks[VUU]{https://arxiv.org/abs/2609.03611}{}{}                                                                                                                                                                                                               & \DatasetCell{Offline}           & \DatasetCell{Failure Detection} & \DatasetCell{Robot Task-Success Judging}  & \DatasetCell{2,197}             & \DatasetCell{Attempts}             & \DatasetCell{\mbox{\DatasetCoverage{III}{E}}} \\
      \SurveyTableRowColor{white}
      \DatasetName{SAFE-VERSE~\cite{CVPR2026AGENTSAFE}}                 & \DatasetLinks[VUU]{https://arxiv.org/abs/2506.14697}{}{}                                                                                                                                                                                                               & \DatasetCell{Simulation}        & \DatasetCell{Task Capability}   & \DatasetCell{Safe Execution}              & \DatasetCell{1,350}             & \DatasetCell{Hazardous Tasks}      & \DatasetCell{\mbox{\DatasetCoverage{III}{E}}} \\
      \SurveyTableRowColor{SurveyTableStripe}
      \DatasetName{IS-Bench~\cite{AAAI2026Isbench}}                     & \DatasetLinks[VVV]{https://arxiv.org/abs/2506.16402}{https://github.com/AI45Lab/IS-Bench}{https://huggingface.co/datasets/Ursulalala/IS_Bench_dataset}                                                                                                                 & \DatasetCell{Simulation}        & \DatasetCell{Task Capability}   & \DatasetCell{Safe Action Ordering}        & \DatasetCell{161}               & \DatasetCell{Eval. Scenarios}      & \DatasetCell{\mbox{\DatasetCoverage{III}{E}}} \\
      \SurveyTableRowColor{white}
      \DatasetName{RecoverBench~\cite{NeurIPS2026RecoverBench}}         & \DatasetLinks[UVV]{}{https://github.com/ErrorRecoveryBenchmark/ErrorRecoveryBenchmark}{https://huggingface.co/datasets/ErrorRecovery/ErrorRecoveryBenchmark}                                                                                                           & \DatasetCell{Simulation}        & \DatasetCell{Task Capability}   & \DatasetCell{Manipulation Recovery}       & \DatasetCell{1,360}             & \DatasetCell{Error Scenes}         & \DatasetCell{\mbox{\DatasetCoverage{III}{E}}} \\
      \SurveyTableRowColor{SurveyTableStripe}
      \DatasetName{FRBench-Sim~\cite{NeurIPS2026RePOVLA}}               & \DatasetLinks[VVU]{https://arxiv.org/abs/2605.09410}{https://github.com/1018weijia/FR_Bench}{}                                                                                                                                                                         & \DatasetCell{Simulation}        & \DatasetCell{Task Capability}   & \DatasetCell{Post-Failure Recovery}       & \DatasetCell{10}                & \DatasetCell{Evaluated Tasks}      & \DatasetCell{\mbox{\DatasetCoverage{III}{E}}} \\
      \SurveyTableRowColor{white}
      \DatasetName{Safety-CHORES~\cite{NeurIPS2025SafeVLA}}             & \DatasetLinks[VVV]{https://proceedings.neurips.cc/paper_files/paper/2025/hash/e185c7be603426028c32ae1003a59d78-Abstract-Conference.html}{https://github.com/PKU-Alignment/SafeVLA}{https://huggingface.co/SafetyEmbodiedAI}                                            & \DatasetCell{Simulation}        & \DatasetCell{Task Capability}   & \DatasetCell{Cumulative Safety Cost}      & \DatasetCell{3}                 & \DatasetCell{Task Families}        & \DatasetCell{\mbox{\DatasetCoverage{III}{E}}} \\
      \SurveyTableBottomRule
    \end{tabular}
    \DatasetLegend{%
      \DatasetCoverage{III}{A}\,Adaptive Perception\quad \DatasetCoverage{III}{B}\,Feedback-Driven Planning\quad
      \DatasetCoverage{III}{C}\,Dynamic Harness Orchestration\quad \DatasetCoverage{III}{D}\,Output and Action Refinement\quad
      \DatasetCoverage{III}{E}\,Execution Monitoring and Control.%
    }{%
      QA = question answering, Envs. = environments, and Eval. = evaluation.%
    }
    \endgroup
    \vspace{-6mm}
  \end{table*}}

\newcommand{\VisualAgentSelfEvolutionBenchmarkTable}{%
  \begin{table*}[t]
    \centering
    \begingroup
    \DatasetTableStyle
    \DatasetTableCaption{Benchmarks and Taxonomy Coverage for Self-Evolution Evaluation.}{tab:visual-agent-self-evolution-benchmarks}
    \vspace{-2mm}
    \begin{tabular}{w{l}{105.5pt}w{c}{36pt}w{l}{65pt}w{l}{78pt}w{l}{92.7pt}w{l}{28.2pt}w{l}{68pt}w{c}{34.6pt}}
      \DatasetTopRule
      \DatasetHeader
      \SurveyTableRowColor{white}
      \DatasetName{Continual Manipulation~\cite{zhu2026can}}        & \DatasetLinks[VVP]{https://arxiv.org/abs/2605.26820v3}{https://github.com/Agentic-Intelligence-Lab/ContinualVLA}{https://github.com/Agentic-Intelligence-Lab/RoboNever}           & \DatasetCell{Physical Robot}       & \DatasetCell{Continual Learning}  & \DatasetCell{Acquisition / Retention}      & \DatasetCell{10}    & \DatasetCell{Robot Tasks}             & \DatasetCell{\mbox{\DatasetCoverage{IV}{A}}} \\
      \SurveyTableRowColor{SurveyTableStripe}
      \DatasetName{CCL Benchmark~\cite{zhou2026benchmarking}}       & \DatasetLinks[VVU]{https://arxiv.org/abs/2609.22055}{https://github.com/object814/Compositional-Continual-Learning-Benchmark}{}                                                   & \DatasetCell{Simulation}           & \DatasetCell{Continual Learning}  & \DatasetCell{Visual Dynamics / Forgetting} & \DatasetCell{6}     & \DatasetCell{Task Suites}             & \DatasetCell{\mbox{\DatasetCoverage{IV}{A}}} \\
      \SurveyTableRowColor{white}
      \DatasetName{LIBERO~\cite{NeurIPS2023LIBERO}}                 & \DatasetLinks[VVV]{https://arxiv.org/abs/2306.03310}{https://github.com/Lifelong-Robot-Learning/LIBERO}{https://libero-project.github.io/datasets}                                & \DatasetCell{Simulation}           & \DatasetCell{Continual Learning}  & \DatasetCell{Manipulation / Forgetting}    & \DatasetCell{130}   & \DatasetCell{Tasks in 4 Suites}       & \DatasetCell{\mbox{\DatasetCoverage{IV}{A}}} \\
      \SurveyTableRowColor{SurveyTableStripe}
      \DatasetName{CHORES~\cite{CoLLAs2022CORA}}                    & \DatasetLinks[VVU]{https://proceedings.mlr.press/v199/powers22b.html}{https://github.com/AGI-Labs/continual_rl}{}                                                                 & \DatasetCell{Simulation}           & \DatasetCell{Continual Learning}  & \DatasetCell{Isolated Forgetting}          & \DatasetCell{4}     & \DatasetCell{Task Sequences}          & \DatasetCell{\mbox{\DatasetCoverage{IV}{A}}} \\
      \SurveyTableRowColor{white}
      \DatasetName{MemGUI-Bench~\cite{liu2026memgui}}               & \DatasetLinks[VVV]{https://arxiv.org/abs/2602.06075}{https://github.com/lgy0404/MemGUI-Bench}{https://huggingface.co/datasets/lgy0404/MemGUI-Bench}                               & \DatasetCell{Mobile GUI}           & \DatasetCell{Memory Capability}   & \DatasetCell{Cross-Session Memory}         & \DatasetCell{128}   & \DatasetCell{GUI Tasks}               & \DatasetCell{\mbox{\DatasetCoverage{IV}{B}}} \\
      \SurveyTableRowColor{SurveyTableStripe}
      \DatasetName{M3-Bench~\cite{ICLR2026Seeing}}                  & \DatasetLinks[VVV]{https://arxiv.org/abs/2508.09736}{https://github.com/ByteDance-Seed/m3-agent}{https://huggingface.co/datasets/ByteDance-Seed/M3-Bench}                         & \DatasetCell{Offline}              & \DatasetCell{Memory Capability}   & \DatasetCell{Audiovisual Recall}           & \DatasetCell{1,020} & \DatasetCell{Videos}                  & \DatasetCell{\mbox{\DatasetCoverage{IV}{B}}} \\
      \SurveyTableRowColor{white}
      \DatasetName{SLVMBench~\cite{NeurIPS2026SLVMBench}}           & \DatasetLinks[VUV]{https://arxiv.org/abs/2607.11312}{}{https://huggingface.co/datasets/tsinghua-ee/SLVMBench}                                                                     & \DatasetCell{Offline}              & \DatasetCell{Memory Capability}   & \DatasetCell{Tutorial Learning / Recall}   & \DatasetCell{2,261} & \DatasetCell{QA Pairs}                & \DatasetCell{\mbox{\DatasetCoverage{IV}{B}}} \\
      \SurveyTableRowColor{SurveyTableStripe}
      \DatasetName{EgoStream~\cite{NeurIPS2026EgoStream}}           & \DatasetLinks[VVU]{https://arxiv.org/abs/2605.31557}{https://github.com/coai-iplab/EgoStream}{}                                                                                   & \DatasetCell{Offline}              & \DatasetCell{Memory Capability}   & \DatasetCell{Answer Obsolescence}          & \DatasetCell{2,250} & \DatasetCell{Base Questions}          & \DatasetCell{\mbox{\DatasetCoverage{IV}{B}}} \\
      \SurveyTableRowColor{white}
      \DatasetName{WorldMemArena~\cite{NeurIPS2026WorldMemArena}}   & \DatasetLinks[VVV]{https://arxiv.org/abs/2605.29341}{https://github.com/UCSB-AI/WorldMemArena}{https://huggingface.co/datasets/LCZZZZ/WorldMemArena}                              & \DatasetCell{Offline}              & \DatasetCell{Memory Capability}   & \DatasetCell{Memory Maintenance / Use}     & \DatasetCell{461}   & \DatasetCell{Multi-Session Tasks}     & \DatasetCell{\mbox{\DatasetCoverage{IV}{B}}} \\
      \SurveyTableRowColor{SurveyTableStripe}
      \DatasetName{Mem-Gallery~\cite{bei2026mem}}                   & \DatasetLinks[VVV]{https://aclanthology.org/2026.acl-long.1892/}{https://github.com/YuanchenBei/Mem-Gallery}{https://huggingface.co/datasets/Ethan-Bei/Mem-Gallery}               & \DatasetCell{Offline}              & \DatasetCell{Memory Capability}   & \DatasetCell{Multimodal Memory Updates}    & \DatasetCell{1,711} & \DatasetCell{All Eval. QA Pairs}      & \DatasetCell{\mbox{\DatasetCoverage{IV}{B}}} \\
      \SurveyTableRowColor{white}
      \DatasetName{MemLeak~\cite{NeurIPS2026MemLeak}}               & \DatasetLinks[VVV]{https://arxiv.org/abs/2606.29788}{https://github.com/kuan-wang/memleak-bench}{https://huggingface.co/datasets/memleak-bench/memleak-benchmark}                 & \DatasetCell{Offline}              & \DatasetCell{Memory Capability}   & \DatasetCell{Deletion Residuals}           & \DatasetCell{300}   & \DatasetCell{Probes / Mode (Main)}    & \DatasetCell{\mbox{\DatasetCoverage{IV}{B}}} \\
      \SurveyTableRowColor{SurveyTableStripe}
      \DatasetName{LMEE-Bench~\cite{CVPR2026Explore}}               & \DatasetLinks[VVV]{https://arxiv.org/abs/2601.10744}{https://github.com/wangsen99/LMEE}{https://huggingface.co/datasets/wangsen99/LMEE-Bench}                                     & \DatasetCell{Simulation}           & \DatasetCell{Memory Capability}   & \DatasetCell{Exploration / Recall}         & \DatasetCell{166}   & \DatasetCell{Test Tasks}              & \DatasetCell{\mbox{\DatasetCoverage{IV}{B}}} \\
      \SurveyTableRowColor{white}
      \DatasetName{EvoNav-Bench~\cite{wang2026evonav}}              & \DatasetLinks[VUU]{https://arxiv.org/abs/2609.08292v2}{}{}                                                                                                                        & \DatasetCell{Simulation}           & \DatasetCell{Experience Reuse}    & \DatasetCell{Reuse / Outdated Scenes}      & \DatasetCell{776}   & \DatasetCell{Test Tasks}              & \DatasetCell{\mbox{\DatasetCoverage{IV}{B}}} \\
      \SurveyTableRowColor{SurveyTableStripe}
      \DatasetName{IR2R-CE~\cite{CVPR2023IterativeVLN}}             & \DatasetLinks[VVU]{https://arxiv.org/abs/2210.03087}{https://github.com/jacobkrantz/IVLN-CE}{}                                                                                    & \DatasetCell{Simulation}           & \DatasetCell{Experience Reuse}    & \DatasetCell{Cross-Episode Map Reuse}      & \DatasetCell{414}   & \DatasetCell{Navigation Tours}        & \DatasetCell{\mbox{\DatasetCoverage{IV}{B}}} \\
      \SurveyTableRowColor{white}
      \DatasetName{EmbodiedCLBench~\cite{huang2026embodiedclbench}} & \DatasetLinks[VVV]{https://arxiv.org/abs/2610.13021}{https://github.com/PKU-VaLuE-Lab/EmbodiedCLBench}{https://huggingface.co/datasets/PKU-VaLuE-Lab/EmbodiedCLBench}             & \DatasetCell{Offline Visual Graph} & \DatasetCell{Skill Acquisition}   & \DatasetCell{Visual Skill Composition}     & \DatasetCell{268}   & \DatasetCell{Task Graphs}             & \DatasetCell{\mbox{\DatasetCoverage{IV}{C}}} \\
      \SurveyTableRowColor{SurveyTableStripe}
      \DatasetName{SEAR~\cite{wang2026sear}}                        & \DatasetLinks[VPU]{https://arxiv.org/abs/2610.18792}{https://github.com/Self-Evolving-Agentic-Robotics/SEAR}{}                                                                    & \DatasetCell{Sim. / Physical}      & \DatasetCell{Experience Reuse}    & \DatasetCell{Skill Reuse / Transfer}       & \DatasetCell{335}   & \DatasetCell{330 Sim. / 5 Real Tasks} & \DatasetCell{\mbox{\DatasetCoverage{IV}{C}}} \\
      \SurveyTableRowColor{white}
      \DatasetName{PhysRSI-eval~\cite{li2026physical}}              & \DatasetLinks[VPU]{https://arxiv.org/abs/2610.26848v1}{https://github.com/PhysRSI-Eval/PhysRSI-Eval-code}{}                                                                       & \DatasetCell{Simulation}           & \DatasetCell{Workspace Evolution} & \DatasetCell{Robot Revision / Transfer}    & \DatasetCell{18}    & \DatasetCell{Robot Tasks}             & \DatasetCell{\mbox{\DatasetCoverage{IV}{D}}} \\
      \SurveyTableRowColor{SurveyTableStripe}
      \DatasetName{OpenRSI Index~\cite{openrsi2026index}}           & \DatasetLinks[VVP]{https://index.openrsi.foundation/}{https://github.com/OpenRSI-Foundation/OpenRSI-Index}{https://huggingface.co/datasets/OpenRSI-Foundation/OpenRSI-Index-Logs} & \DatasetCell{General RSI Suite}    & \DatasetCell{Inference Harness}   & \DatasetCell{Web Context / Video Pointing} & \DatasetCell{2}     & \DatasetCell{Named Visual Tasks}      & \DatasetCell{\mbox{\DatasetCoverage{IV}{D}}} \\
      \SurveyTableRowColor{white}
      \DatasetName{SelfImprove-Bench~\cite{xu2026selfimprovebench}} & \DatasetLinks[VVU]{https://arxiv.org/abs/2611.02797}{https://github.com/theseus-labs-rsi/SelfImprove-Bench}{}                                                                     & \DatasetCell{Minecraft / Astray}   & \DatasetCell{Inference Harness}   & \DatasetCell{Cross-Game Transfer}          & \DatasetCell{10}    & \DatasetCell{Fixed Tasks (5 / Game)}  & \DatasetCell{\mbox{\DatasetCoverage{IV}{D}}} \\
      \SurveyTableRowColor{SurveyTableStripe}
      \DatasetName{Procgen Curriculum Suite~\cite{lee2026curate}}   & \DatasetLinks[VVU]{https://openreview.net/forum?id=DlnvWfoIgv}{https://github.com/montrealrobotics/curate-code}{}                                                                 & \DatasetCell{Visual Games}         & \DatasetCell{Curriculum Learning} & \DatasetCell{Adaptive Task Sampling}       & \DatasetCell{16}    & \DatasetCell{Games}                   & \DatasetCell{\mbox{\DatasetCoverage{IV}{E}}} \\
      \SurveyTableBottomRule
    \end{tabular}
    \DatasetLegend{%
      \DatasetCoverage{IV}{A}\,Model-Level Evolution\quad \DatasetCoverage{IV}{B}\,Memory-Driven Capability Evolution\quad
      \DatasetCoverage{IV}{C}\,Skill and Tool-Level Evolution\quad \DatasetCoverage{IV}{D}\,Orchestration-Level Evolution\quad
      \DatasetCoverage{IV}{E}\,Dynamic Data and Task Evolution.%
    }{%
      CCL Benchmark = Compositional Continual Learning Benchmark. Eval. = evaluation, QA = question answering, and Sim. = simulation.%
    }
    \endgroup
    \vspace{-6mm}
  \end{table*}}

\newcommand{\VisualAgentRSIBenchmarkTable}{%
  \begin{table*}[t]
    \centering
    \begingroup
    \DatasetTableStyle

    \DatasetTableCaption{Benchmarks and Taxonomy Coverage for Evaluating RSI.}{tab:visual-agent-rsi-benchmarks}
    \vspace{-2mm}
    \begin{tabular}{w{l}{105.5pt}w{c}{36pt}w{l}{65pt}w{l}{78pt}w{l}{92.7pt}w{l}{28.2pt}w{l}{68pt}w{c}{34.6pt}}
      \DatasetTopRule
      \DatasetHeader
      \SurveyTableRowColor{white}
      \DatasetName{Flex-RSI~\cite{zhang2026flex}}                      & \DatasetLinks[VPP]{https://arxiv.org/abs/2611.08946v1}{https://github.com/flex-rsi/rsi-bench}{https://huggingface.co/datasets/Flex-RSI/RSI-Dataset} & \DatasetCell{Vision / Robotics} & \DatasetCell{Harness RSI}          & \DatasetCell{Recursive Harness Updates}  & \DatasetCell{Unlisted} & \DatasetCell{Selected Tasks}     & \DatasetCell{\mbox{\DatasetBadge{DatasetLevelVFill}{DatasetLevelVText}{V}}} \\
      \SurveyTableRowColor{SurveyTableStripe}
      \DatasetName{AI4AI-Bench~\cite{chi2026ai4ai}}                    & \DatasetLinks[VVU]{https://arxiv.org/abs/2608.20318v1}{https://github.com/Einsia/AI4AI-Bench}{}                                                     & \DatasetCell{General RSI Suite} & \DatasetCell{Update-Method Design} & \DatasetCell{Diffusion RL / CLIP Merge}  & \DatasetCell{2}        & \DatasetCell{Visual Families}    & \DatasetCell{\mbox{\DatasetBadge{DatasetLevelVFill}{DatasetLevelVText}{V}}} \\
      \SurveyTableRowColor{white}
      \DatasetName{RSI-Exam~\cite{rsi-exam-2026}}                      & \DatasetLinks[VPP]{https://rsi-exam.ai/}{https://github.com/aiming-lab/RSI-Exam}{https://huggingface.co/datasets/RSI-Exam/RSI-Exam}                 & \DatasetCell{General RSI Suite} & \DatasetCell{Update-Method Design} & \DatasetCell{VLM Reward / Training}      & \DatasetCell{1}        & \DatasetCell{Counting Task}      & \DatasetCell{\mbox{\DatasetBadge{DatasetLevelVFill}{DatasetLevelVText}{V}}} \\
      \SurveyTableRowColor{SurveyTableStripe}
      \DatasetName{AutoResearchExam~\cite{ramesh2026autoresearchexam}} & \DatasetLinks[VVU]{https://benchmarks.bespokelabs.ai/autoresearchexam/}{https://github.com/bespokelabsai/AutoResearchExam}{}                        & \DatasetCell{Research Suite}    & \DatasetCell{Procedure Design}     & \DatasetCell{CLIP Merge / Video Caching} & \DatasetCell{2}        & \DatasetCell{Named Visual Tasks} & \DatasetCell{\mbox{\DatasetBadge{DatasetLevelVFill}{DatasetLevelVText}{V}}} \\
      \SurveyTableRowColor{white}
      \DatasetName{Vals RSI Index~\cite{valsai2026rsiindex}}           & \DatasetLinks[VUU]{https://vals-ai.com/benchmarks/rsi_index}{}{}                                                                                    & \DatasetCell{General RSI Suite} & \DatasetCell{Evaluator Design}     & \DatasetCell{Visual Web-App Judge}       & \DatasetCell{24}       & \DatasetCell{Held-Out Apps}      & \DatasetCell{\mbox{\DatasetBadge{DatasetLevelVFill}{DatasetLevelVText}{V}}} \\
      \SurveyTableBottomRule
    \end{tabular}
    \DatasetLegend{%
      \DatasetBadge{DatasetLevelVFill}{DatasetLevelVText}{V}\,RSI and Mechanism Design.%
    }{%
      RL = reinforcement learning and VLM = vision-language model.%
    }
    \endgroup
    \vspace{-6mm}
  \end{table*}}

A \emph{visual agent}'s task results show what it can do under specified conditions. To evaluate how it improves, we follow those results across attempts and ask what changes the agent carries forward and how revisions to its improvement process affect later updates.

\subsection{Performance Evaluation at the Task-Level}
\label{sec:benchmark-understanding}

Task-level evaluation measures goal attainment and output quality per attempt. It also examines how the agent uses visual evidence, executes actions, recovers from errors, and respects constraints. Table~\ref{tab:visual-agent-task-benchmarks} lists representative benchmarks.

\VisualAgentTaskBenchmarkTable

\noindent$\bullet$\ \textbf{Perception and Grounded Reasoning.} A grounded response depends on inferring the relevant task state from visual evidence. Offline question-answering benchmarks probe visually grounded state understanding and reasoning about actions or changes over time~\cite{gupta2026agentvqa,fu2026video,ICLR2026ENACT,ACL2026MPRGUI}. For active evidence gathering, ActiView~\cite{ACL2025ActiView} measures answer accuracy after the agent selects image regions to inspect, while the active track of OpenEQA~\cite{CVPR2024OpenEQA} also accounts for scene exploration steps. ActiveArena-Bench~\cite{li2026activearena} tests how viewpoint control and interaction help agents acquire evidence for manipulation.

\noindent$\bullet$\ \textbf{Planning, Execution, and Tool Interaction.} Once an agent has a goal and evidence about its current state, evaluation follows whether its decisions carry the task through to completion. VisualWebArena~\cite{ACL2024VisualWebArena} uses execution-based checks to score tasks that require interpreting webpage images, while MemU-Bench~\cite{liu2026memubench} measures whether agents complete mobile tasks whose instructions require context from a user's earlier screenshots and actions. Other GUI and embodied benchmarks check the resulting state or apply task-specific success criteria~\cite{yuan2026osworld2,NeurIPS2026WebArenaPro,ACL2026NaturalGAIA,CVPR2026ProSoftArena,CVPR2026Ego2Web,ACL2026GUI0}. ALFRED~\cite{CVPR2020ALFRED} reports both overall household task success and the fraction of goal conditions met. Other long-horizon benchmarks track milestones along a trajectory or test how agents adapt to changes during execution~\cite{tupini2026asgardbench,AAAI2026DeepPHY,karten2026pokeagent}. ProBench~\cite{AAAI2026Probench} also checks whether required actions occurred when the final GUI state alone cannot establish success. Tool-centered evaluations pair task outcomes with analysis of tool selection and use~\cite{NeurIPS2026AgentVista,liu2026egobench,ICLR2026OSWorldMCP}.

\noindent$\bullet$\ \textbf{Artifact and Output Quality.} When the task produces an artifact, its quality depends on how well it meets the user's requirements. Benchmarks for slides, webpages, and images assess content, appearance, or preservation of requested elements through rubrics and judgments of the resulting artifacts~\cite{ICML2026PPTEval,wang2026refcraft,he2026evaluation,NeurIPS2026GenEvolve}. ComfyBench~\cite{CVPR2025ComfyBench} checks whether an agent-designed workflow runs and whether its image or video satisfies the request. In VisualAgentBench~\cite{ICLR2025VisualAgentBench}, VAB-CSS scores repairs by the structural similarity of the final webpage rendering to its target screenshot. For 3D generation, 3DCodeBench~\cite{NeurIPS20263DCodeBench} scores objects built by executing code, while 3DHarnessBench~\cite{liu20263dharnessbench} compares reconstructions from one or more fixed views, agent-selected views, or direct queries to the target geometry.

\noindent$\bullet$\ \textbf{Reliability and Recovery.} When an error interrupts execution, the agent must recognize the problem and find a way to resume progress. FailBench~\cite{navasardyan2026failbench} measures failure detection from recorded attempts, while CUA-Recovery~\cite{hu2026cuarecovery} and RecoverBench~\cite{NeurIPS2026RecoverBench} assess task completion from error states. FRBench-Sim~\cite{NeurIPS2026RePOVLA} compares manipulation success with and without a grasp disturbance during execution.

\noindent$\bullet$\ \textbf{Safety and Security.} Whether an agent completes a task safely depends on the actions it takes and how it responds to misleading inputs. Safety benchmarks use rule- or state-based checks to assess behavioral constraints and whether mitigation occurs before unsafe actions~\cite{CVPR2026AGENTSAFE,AAAI2026Isbench}. Safety-CHORES~\cite{NeurIPS2025SafeVLA} records both task success and the cumulative cost of safety violations as agents navigate and manipulate objects using camera observations. VPI-Bench~\cite{ICLR2026VPIBench} tests robustness to prompt injection through visual inputs.

\subsection{Evaluation of Self-Evolution}
\label{sec:benchmark-experience}

\VisualAgentSelfEvolutionBenchmarkTable

Self-evolution evaluation asks what the agent gains on target tasks after a retained model or harness update, what it retains during continued learning, and what transfers to new tasks or conditions. Table~\ref{tab:visual-agent-self-evolution-benchmarks} lists representative benchmarks.

\noindent$\bullet$\ \textbf{Capability Acquisition.} Acquisition can be assessed by comparing performance on the target tasks before and after the agent incorporates new experience. A control with a matched budget and the same evaluation tasks helps isolate the contribution of the update. Continual manipulation protocols track task success as the agent learns from demonstrations for newly introduced tasks~\cite{zhu2026can}. WorldMemArena~\cite{NeurIPS2026WorldMemArena} checks how new observations inform later responses across sessions. On MemGUI-Bench~\cite{liu2026memgui}, agents retry mobile tasks from the same device state, allowing task success to be compared with and without memory from earlier attempts. To isolate the effect of how demonstrations are collected, RECALL~\cite{karli2026recall} compares success on LIBERO-10~\cite{NeurIPS2023LIBERO} after fine-tuning with recovery demonstrations from high-uncertainty states or with demonstrations from task starts, holding the number of new demonstrations per task fixed. The Procgen Curriculum Suite~\cite{lee2026curate} compares feedback-driven and random curricula for pixel-based game agents, measuring target-task success and the number of training updates needed to reach a performance threshold.

\noindent$\bullet$\ \textbf{Capability Retention.} Retesting earlier tasks during subsequent learning reveals how an update affects previously acquired capabilities. LIBERO~\cite{NeurIPS2023LIBERO} quantifies forgetting from the drop in a visual policy's success on previously learned manipulation tasks. In the CHORES benchmark of CORA~\cite{CoLLAs2022CORA}, changes in return on earlier visually guided household tasks show how well a policy preserves them as training continues. Related protocols track changes in manipulation success or visual prediction over a task sequence~\cite{zhu2026can,zhou2026benchmarking}. For memory, the expected answer can change with the world, so WorldMemArena and EgoStream test whether agents keep their answers current as observations become outdated~\cite{NeurIPS2026WorldMemArena,NeurIPS2026EgoStream}. MemLeak~\cite{NeurIPS2026MemLeak} probes whether residual evidence still reveals information marked for deletion.

\noindent$\bullet$\ \textbf{Generalization and Transfer.} Held-out tasks and changed conditions test how far a retained update generalizes. EmbodiedCLBench~\cite{huang2026embodiedclbench} evaluates whether skills learned on simpler tasks support unseen compositions. PhysRSI-eval~\cite{li2026physical} tests retained workspaces on new manipulation tasks, while SEAR~\cite{wang2026sear} compares skill and memory reuse against runs without a library and time-matched restarts. In SelfImprove-Bench~\cite{xu2026selfimprovebench}, the screenshot-based Minecraft and Astray tasks test whether a revised harness transfers to unseen instances of the same game or to the other game. Retained video harnesses are tested on held-out question answering~\cite{xu2026videoharness,luo2026video}, and the MolmoWeb and Molmo2 tasks in OpenRSI Index~\cite{openrsi2026index} assess revised inference strategies on held-out cases. IR2R-CE~\cite{CVPR2023IterativeVLN} tests whether maps retained across routes help agents follow new instructions in the same building. Other benchmarks probe memory reuse in visual reasoning and navigation~\cite{ICLR2026Seeing,NeurIPS2026WorldMemArena,CVPR2026Explore,wang2026evonav}. SLVMBench~\cite{NeurIPS2026SLVMBench} tests transfer from supplied video tutorials to new questions, while Mem-Gallery~\cite{bei2026mem} probes memory using multimodal conversation histories.

\subsection{Evaluation of Recursive Self-Improvement}
\label{sec:benchmark-rsi}

\VisualAgentRSIBenchmarkTable

Self-evolution evaluation measures capability changes after retained updates from a fixed updater. At the next level, recursive self-improvement (RSI) evaluation asks whether revising the updater improves later updates. Table~\ref{tab:visual-agent-rsi-benchmarks} covers multi-round tests of reuse and mechanism-design tasks.

\noindent$\bullet$\ \textbf{RSI Benchmarks for Visual Agents.} Successive evaluation rounds can show where an adopted revision takes effect in the next update. Flex-RSI~\cite{zhang2026flex} evaluates visual and robotics tasks over repeated executor and improver rounds, allowing the improver to revise both the executor and its own procedures for later reuse. Validation checks candidate solutions, while held-out tests measure performance under fixed budgets.

\noindent$\bullet$\ \textbf{Mechanism-Design Tasks with Visual Inputs.} Each general-purpose research benchmark includes visual tasks as one subset of its design problems. Scores on these visual tasks can assess the outcome of a mechanism revision, while evidence of visual RSI comes from tracing the revised mechanism to a later retained update. AI4AI-Bench~\cite{chi2026ai4ai} scores redesigned diffusion training and CLIP merging by image aesthetics and ImageNet-V2 accuracy, respectively, and RSI-Exam~\cite{rsi-exam-2026} evaluates a model trained with a redesigned reward on hidden visual counting tasks. AutoResearchExam~\cite{ramesh2026autoresearchexam} tests CLIP merges on held-out classification and scores video caching by fidelity to uncached outputs under a compute limit. Vals RSI Index~\cite{valsai2026rsiindex} measures agreement between a redesigned judge and reference ratings on held-out web apps.

\noindent$\bullet$\ \textbf{Downstream Visual Benchmarks.} Standard visual benchmarks score the agent's performance on a defined set of tasks. In a visual RSI study, testing the initial and revised agents under the same conditions measures downstream gains, while records of a later retained update show whether the revised updater was actually used. VideoEvoHarness~\cite{xu2026videoevoharness} compares multiple-choice accuracy under initial and evolved harnesses on LongVideoBench~\cite{NeurIPS2024LongVideoBench}, MMVU~\cite{CVPR2025MMVU}, and Video-Holmes~\cite{ECCV2026VideoHolmes}, and tracks gains across rounds on a fixed question set. For image generation, RSIgen~\cite{liu2026rsigen} compares instruction adherence and image quality on held-out requests drawn from SearchGen~\cite{wang2026search}. AutoMMemo~\cite{sui2026learning} evaluates revised memory programs on web and multimodal question-answering tasks, and EmbodiedRSI~\cite{song2026embodiedrsi} measures success on unseen manipulation tasks in RoboCasa365~\cite{ICLR2026RoboCasa365}. JarvisEvo~\cite{CVPR2026Jarvisevo} uses ArtEdit-Bench-Lr to score instruction adherence and fidelity to reference edits after evaluator-guided training. RewardHarness~\cite{zhang2026rewardharness} also assesses image edits, while CineEvolve~\cite{zhang2026cineevolve} scores videos from its evolved production workflow. VeriCUA~\cite{wang2026vericua} measures GUI task success after training an actor with the revised verifier, and SkillRubric~\cite{jiang2026skillrubric} scores a policy trained with revised rubrics on held-out multimodal tool-use tasks.

\section{Open Challenges and Future Directions}
\label{sec:future-directions}

\noindent$\bullet$\ \textbf{Balancing Capability Retention and Generalization.} A central goal of visual agent evolution is to retain useful visual representations, object–action associations, and task execution strategies while generalizing to new visual observations, object instances, spatial configurations, and task objectives. Prior work has explored replay-based learning, knowledge preservation, and adaptation through new task demonstrations to mitigate forgetting and acquire new capabilities~\cite{added_liu2026continualvla,tharwat2026dream,raghavan2026does,zhu2026can}. Relying heavily on past experience can constrain adaptation as conditions change, and updating for new tasks can erode previously acquired capabilities. An open challenge is to preserve transferable visual knowledge across successive updates while selectively adapting task-specific behaviors. Recording the visual and task conditions under which each consolidated skill applies could help agents retrieve, adapt, and recombine relevant skills as observations and objectives change, subject to practical interaction and memory budgets.

\noindent$\bullet$\ \textbf{Preventing Objective Hacking in Visual Agent Evolution.} Evaluating visual agent evolution requires distinguishing capability gains from apparent performance improvements due to exploitation of evaluation objectives, execution environments, or tool interfaces. Sandboxed execution, phase-specific write restrictions, and independent release checks have been explored as ways to constrain agent modifications and detect unreliable improvements~\cite{ICLR2026Darwin,zhang2026flex,wang2026regenharness}. Objective hacking remains possible under execution isolation. In one sandboxed system, high evaluation scores masked persistent tool-use hallucinations~\cite{ICLR2026Darwin}. Even with write restrictions and release checks, agents may exploit evaluation criteria or reuse approved workflows beyond their intended task scope. Maintaining independent evaluation and enforcing task-specific authorization remain challenging throughout successive evolution cycles and workflow reuse. Future safeguards could establish security boundaries outside the agent-editable workflow, independently audit agent behaviors and evaluation outcomes, and incorporate human oversight for consequential deployments.

\noindent$\bullet$\ \textbf{Maintaining Consistency in Long-Horizon Execution.} Long-horizon visual agents must maintain consistency between task objectives, intermediate decisions, and evolving environmental states across successive interactions. Perception errors and unexpected action outcomes can invalidate state assumptions. Acting on these assumptions can compound errors and divert execution from the task objective. Existing methods use visual observations to verify expected action effects and check whether task constraints still hold~\cite{ACL2026VeriGUI,yang2026task}. Incomplete observations and changes in the environment make it difficult to infer overall task progress from these checks. Tracking the observations that support each task milestone could help agents reassess progress when new evidence invalidates earlier state assumptions. Agents could then update their state estimates and revise the remaining plan in line with the original task objective.

\noindent$\bullet$\ \textbf{Verifying the Mechanisms of Recursive Self-Improvement.} Evidence for recursive self-improvement (RSI) depends on establishing whether improvements across successive evolution rounds result from an improved updater or merely from accumulating experience and additional computational resources. Some studies have revised components such as curriculum generators and evaluators and reused them in subsequent training rounds~\cite{ACL2026GUI0,CVPR2026Jarvisevo}. However, showing that an updated component participates in later evolution does not establish that it improves the update process itself, as observed gains may also arise from repeated optimization. Whether updater improvements yield sustained benefits across successive updates remains unclear. Controlled comparisons between revised and original updaters under matched experience and resource budgets could help establish causal contributions, while tracing how revisions alter subsequent data generation, feedback, and learning dynamics could reveal how recursive improvement occurs.

\noindent$\bullet$\ \textbf{Benchmarking Development for Visual Agents.} Task-level metrics alone cannot reveal whether an agent succeeds through reliable perception, planning, and execution or through compensating behaviors and environmental shortcuts. Evaluation protocols have addressed process-level verification of required actions, controlled evaluation of retained workspaces, and multi-round assessment of successive agent revisions~\cite{AAAI2026Probench,li2026physical,zhang2026flex,meng2026agent}. However, checking intermediate actions as additional task outputs does not establish the effectiveness of individual agent components or their contributions to overall performance. Future benchmarks should jointly evaluate task outcomes, component-level capabilities, and the functional contributions of harness components under comparable conditions across successive updates. A shared evaluation framework could combine task-based assessments with component-specific tests and controlled interventions, enabling systematic comparisons of capability development, component effectiveness, reliability, and resource costs across visual agents.

\section{Conclusion}
\label{sec:conclusion}

This survey traces the development of visual agents and provides a systematic overview of recent advances. We first define visual agents and describe their foundations, including the complementary roles of models and agent harnesses. Building on this perspective, we review representative methods in visual understanding, visual generation and editing, computer use, and embodied interaction, using a taxonomy that distinguishes within-task adaptation, self-evolution, and work toward recursive self-improvement (RSI). We then summarize benchmarks, metrics, and protocols for assessing task performance and reliability, tracking capability acquisition, retention, and transfer, and evaluating evidence for RSI. Finally, we discuss open challenges and outline directions for developing visual agents that improve reliably and generalize across tasks and environments.

\bibliographystyle{IEEEtran}
\bibliography{bibliography-control,references_conference, references_preprint_less, references_url}
\end{document}